\pdfoutput=1

\documentclass[11pt]{article}

\usepackage{ACL2023}

\usepackage{times}
\usepackage{latexsym}

\usepackage[T1]{fontenc}

\usepackage[utf8]{inputenc}

\usepackage{microtype}

\usepackage{inconsolata}

\usepackage{multirow}
\usepackage{array}
\usepackage{graphicx}
\usepackage{epstopdf}
\usepackage{amsmath}
\usepackage{amssymb}

\usepackage{booktabs}

\newcommand*{\circled}[1]{\lower.5ex\hbox{\tikz\draw (0pt, 0pt)%
    circle (.45em)
    (0pt, 0pt)%
    circle (.38em) node {\makebox[1em][c]{\scriptsize #1}};}}

\usepackage[normalem]{ulem}

\usepackage[inline]{enumitem}
\usepackage{romannum}
\AtBeginDocument{\pagenumbering{arabic}}

\usepackage{float}

\usepackage{mathtools}

\usepackage{cleveref}
\usepackage{hyperref}

\usepackage{longtable}

\usepackage{algorithm}
\usepackage{algorithmic}

\newcommand{\tabincell}[2]{
\begin{tabular}{@{}#1@{}}#2\end{tabular}
}

\usepackage{color}

\definecolor{lgreen}{RGB}{0,204,102}
\definecolor{lred}{RGB}{255,51,51}
\definecolor{dred}{RGB}{173,0,0}
\definecolor{blue}{RGB}{102,178,255}
\definecolor{purple}{RGB}{204,0,204}

\definecolor{syellow}{RGB}{204,204,0}
\definecolor{sgreen}{RGB}{0,153,77}
\definecolor{sblue}{RGB}{0,102,204}
\definecolor{sred}{RGB}{204,0,0}
\definecolor{spurple}{RGB}{204,0,204}
\definecolor{smediumvioletred}{RGB}{199,21,133}
\definecolor{Intersection}{RGB}{255,51,51}
\definecolor{Optimal}{RGB}{153,0,153}
\definecolor{HoneyOrange}{RGB}{255,179,102}
\definecolor{darkred}{RGB}{139,0,0}

\usepackage{pgfplots}
\usepackage{tikz}
\usetikzlibrary{3d}

\usepackage{abstract}

\usepackage{lipsum}

\title{Controllable Text Generation via Probability Density Estimation\\ in the Latent Space}

\author{Yuxuan Gu$^{\dag}$,Xiaocheng Feng$^{\dag \ddag}$,Sicheng Ma$^{\dag}$,Lingyuan Zhang$^{\dag}$,Heng Gong$^{\dag}$,Weihong Zhong$^{\dag}$,Bing Qin$^{\dag \ddag}$\\
  $^{\dag}$Harbin Institute of Technology\quad \quad \quad $^\ddag$ Peng Cheng Laboratory\\
  \texttt{\{yxgu,xcfeng,scma,lyzhang,hgong,whzhong,qinb\}@ir.hit.edu.cn} \\}

\begin{document}
\maketitle

\renewcommand{\abstracttextfont}{\fontsize{10pt}{11.5pt}\selectfont}
\setlength{\absleftindent}{0.6cm}
\setlength{\absrightindent}{0.6cm}
\begin{abstract}

\noindent Previous work on controllable text generation has explored the idea of control from the latent space, such as optimizing a representation with attribute-specific classifiers or sampling one from relevant discrete points. However, they cannot effectively model a complex space with diverse attributes, high dimensionality, and asymmetric structure, leaving subsequent controls unsatisfying.
In this work, we propose a novel control framework using probability density estimation in the latent space. Our method utilizes an invertible transformation function, the Normalizing Flow, that maps the complex distributions in the latent space to simple Gaussian distributions in the prior space. Thus, we can perform sophisticated and flexible controls in the prior space and feed the control effects back into the latent space owing to the 
bijection property of invertible transformations. Experiments on single-attribute and multi-attribute control reveal that our method outperforms several strong baselines on attribute relevance and text quality, achieving a new SOTA. Further analysis of control strength adjustment demonstrates the flexibility of our control strategy\footnote{\url{https://github.com/HappyGu0524/MultiControl}.}.
\end{abstract}

\section{Introduction}

Controllable text generation, a fundamental issue in natural language generation, refers to generating fluent and attractive sentences conditioned on target attributes \cite{zhang2022survey}.
With the development of pre-trained language models \cite{zhao2023survey}, early work explores converting generative language models to conditional models by altering their parameters via fine-tuning \cite{ziegler2019finetuning, keskarCTRL2019} or reinforcement learning \cite{khalifa2020distributional}.
Due to the high cost of modifying parameters \cite{NEURIPS2020_1457c0d6, zhang2022opt}, current control approaches prefer leaving pre-trained language models fixed \cite{Dathathri2020Plug, krause-etal-2021-gedi-generative}.

\begin{figure}[t]
  \centering
  \includegraphics[width=\columnwidth]{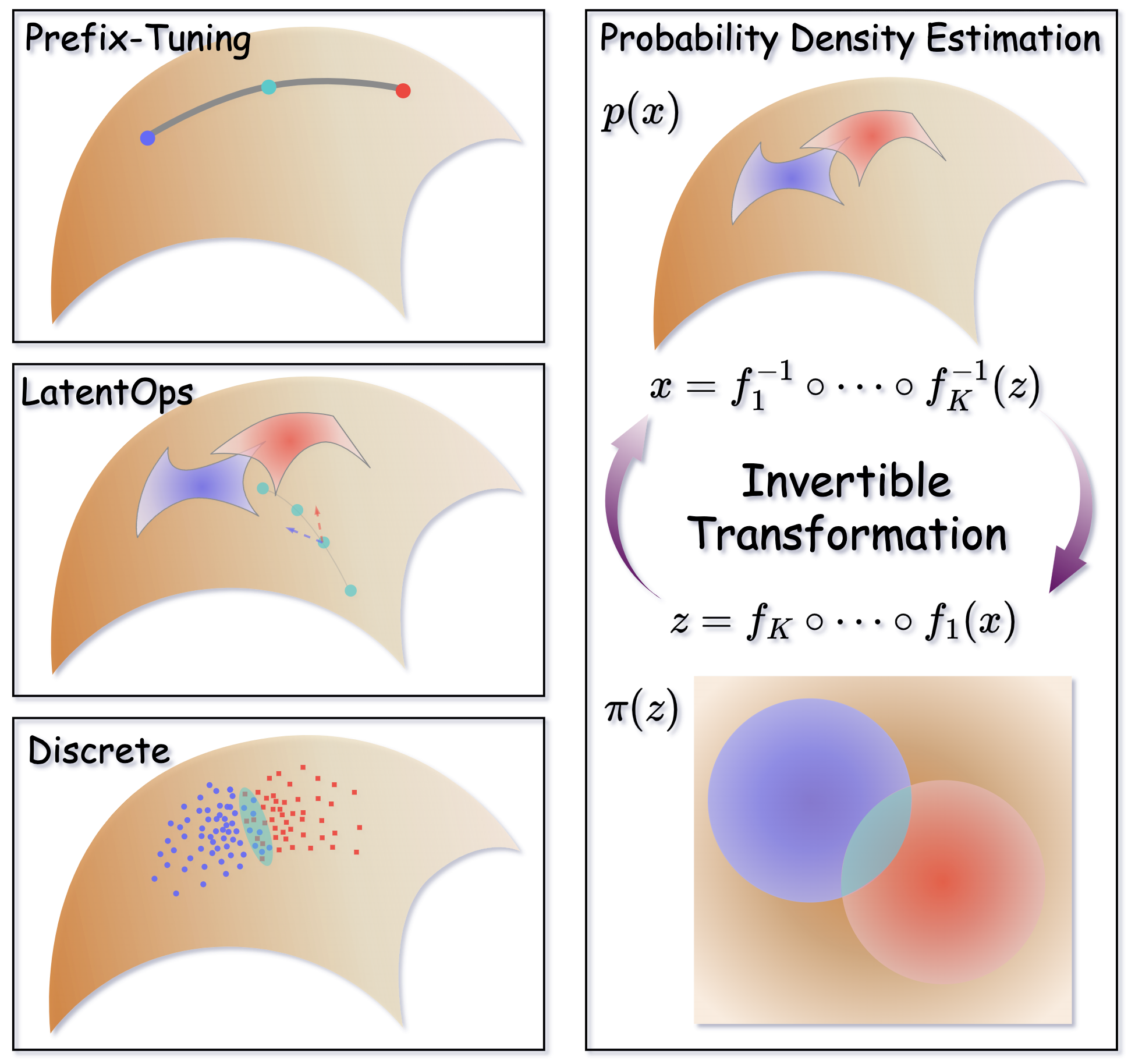}
  \caption{Illustration of methods controlling in Latent Space. \textcolor{HoneyOrange}{Orange} background denotes the latent space. \textcolor{blue}{Blue} and \textcolor{lred}{red} represent two attributes. \textbf{Prefix-Tuning} represents attributes with points in latent space and composes them by interpolation. \textbf{LatentOps} uses classifiers to estimate continuous distributions of attributes and control by optimizing in latent space. \textbf{Discrete} maps sentences to discrete samples in latent space and controls with direct searching. \textbf{Our method} deploys probability density estimation by transforming the complex latent space into a well-formed prior space, where common control strategies can be more effective. See \S \ref{app:ComplexSpace} for more details of the latent space's defects.}
  \label{fig:1}
\end{figure}
Recent studies perform impressive control by influencing the fixed language model from the latent space \cite{yu-etal-2021-attribute-alignment, qian-etal-2022-controllable} with prefix-tuning \cite{li-liang-2021-prefix}.
However, inefficient and unreliable modeling of the complex latent space remains a problem that plagues control performance. As shown in the left part of Figure \ref{fig:1}, \citet{gu2022distributional} reveal that distributions of attributes in high dimensional latent space are usually asymmetric and even non-convex, making simple control strategies inefficient, including interpolation methods like Prefix-Tuning \cite{qian-etal-2022-controllable} and optimization approaches like LatentOps \cite{liu2022composable}.
For example, interpolation may exceed the support set of distributions, making generated sentences unable to acquire desired attributes.
Besides, the optimization process can stuck in the saddle or local optimal points.
Although mitigating the problem by discrete modeling and direct searching, Discrete \cite{gu2022distributional} introduces a more complicated control process, where searching for the intersection of attributes is vulnerable to the high-dimensionality of space and noise in samples.

In this paper, we alleviate problems above by better modeling the latent space.
As shown in the right of Figure \ref{fig:1}, we propose probability density estimation in latent space by invertible transformation, where complex distributions of attributes in latent space are mapped (\textit{bijection between continuous spaces}) to simple ones, such as Gaussian distributions, in prior space. 
Thus, traditional control strategies such as interpolation can be tractable and explainable in this normalized prior space.
In the inference stage, our paradigm becomes: control attributes in prior space, then activate the language model in latent space.
Furthermore, we explore the relationship between the latent space and our prior space and attempt to prove under what circumstances the control in the prior space can be effectively fed back into the latent space.

We conduct experiments on single-attribute control and multi-attribute control.
Datasets we use are IMDb movie reviews \cite{maas-etal-2011-learning} for Sentiment, AGNews \cite{NIPS2015_250cf8b5} for Topic, and Jigsaw Toxic Comment Classification Challenge Dataset for Detoxification.
We measure the control ability of our method using the correlation of generated sentences with each attribute.
For text generation quality, we evaluate sentences with perplexity and distinctness concerning fluency and diversity.
Results show that our method can significantly outperform baseline models and analytical experiments on control strength adjustment reveal our flexibility.
The main contributions of our work are summarized as follows:
\begin{itemize}[topsep=2pt,parsep=0pt]
    \item We propose a novel framework that introduces a well-formed prior space for effective and flexible control via invertible transformation.
    \item We theoretically explore approaches to exploit invertibility to feed control in the prior space back into the latent space.
    \item We experimentally reveal the effectiveness of our method compared to previous SOTA.
\end{itemize}

\section{Related Work}
\subsection{Controllable Text Generation}
Variational autoencoders are often used for controllable text generation \cite{10.5555/3305381.3305545, duan-etal-2020-pre, mai-etal-2020-plug} before the prosperity of large-scale pre-trained language models \cite{radford2019language}. 
Traditional control approaches like fine-tuning \cite{ficler-goldberg-2017-controlling, ziegler2019finetuning, keskarCTRL2019} and reinforcement learning \cite{khalifa2020distributional} gradually become infeasible with the rapid increase of language models' parameters.
Recent methods investigate control with fixed language models, including biasing the token distribution during decoding \cite{Dathathri2020Plug, krause-etal-2021-gedi-generative, yang-klein-2021-fudge, liu-etal-2021-dexperts, gu-etal-2022-improving, meng2022controllable}, optimization in the language space \cite{kumar2021controlled, qin2022cold, mireshghallah-etal-2022-mix, kumar2022constrained}, and optimization in the latent space \cite{yu-etal-2021-attribute-alignment, qian-etal-2022-controllable, carlsson-etal-2022-fine, yang2022tailor, liu2022composable, lu2022quark, zhang2022discup, gu2022distributional}. Another work trains a denoising diffusion language model before controlling sentence attributes in the denoising process \cite{li2022diffusion}.

\subsection{Normalizing Flow}
The Normalizing Flow \cite{dinh2014nice, dinh2016density, NEURIPS2018_d139db6a, NIPS2016_ddeebdee, NIPS2017_6c1da886}, consisting of a sequence of invertible transformations for continuous variables, is a powerful deep generative model \cite{kingma2013auto, goodfellow2020generative, NEURIPS2020_4c5bcfec} that enables capturing the inner probabilistic distribution of complex and high-dimensional data \cite{8354080}, including images and text. In natural language processing, Normalizing Flows are often used as enhanced prior distributions in VAE structures \cite{ma-etal-2019-flowseq, ding-gimpel-2021-flowprior} or as deep generative language models \cite{NEURIPS2019_e046ede6, pmlr-v97-ziegler19a, tang-etal-2021-continuous}.
Besides, \citet{wu2022generative} uses the Normalizing Flow as prefix-tuning for controllable image generation.
However, previous work usually treats Normalizing Flow as an ordinary generative model, easily replaced by stronger models like the denoising diffusion model \cite{NEURIPS2020_4c5bcfec}, while ignoring its invertible property. In this work, we will explore the potential for the flexible application of the Normalizing Flow's invertible feature in controllable text generation.


\begin{figure*}[ht]
  \centering

  \resizebox{2\columnwidth}{!}{
	\includegraphics[scale=0.55]{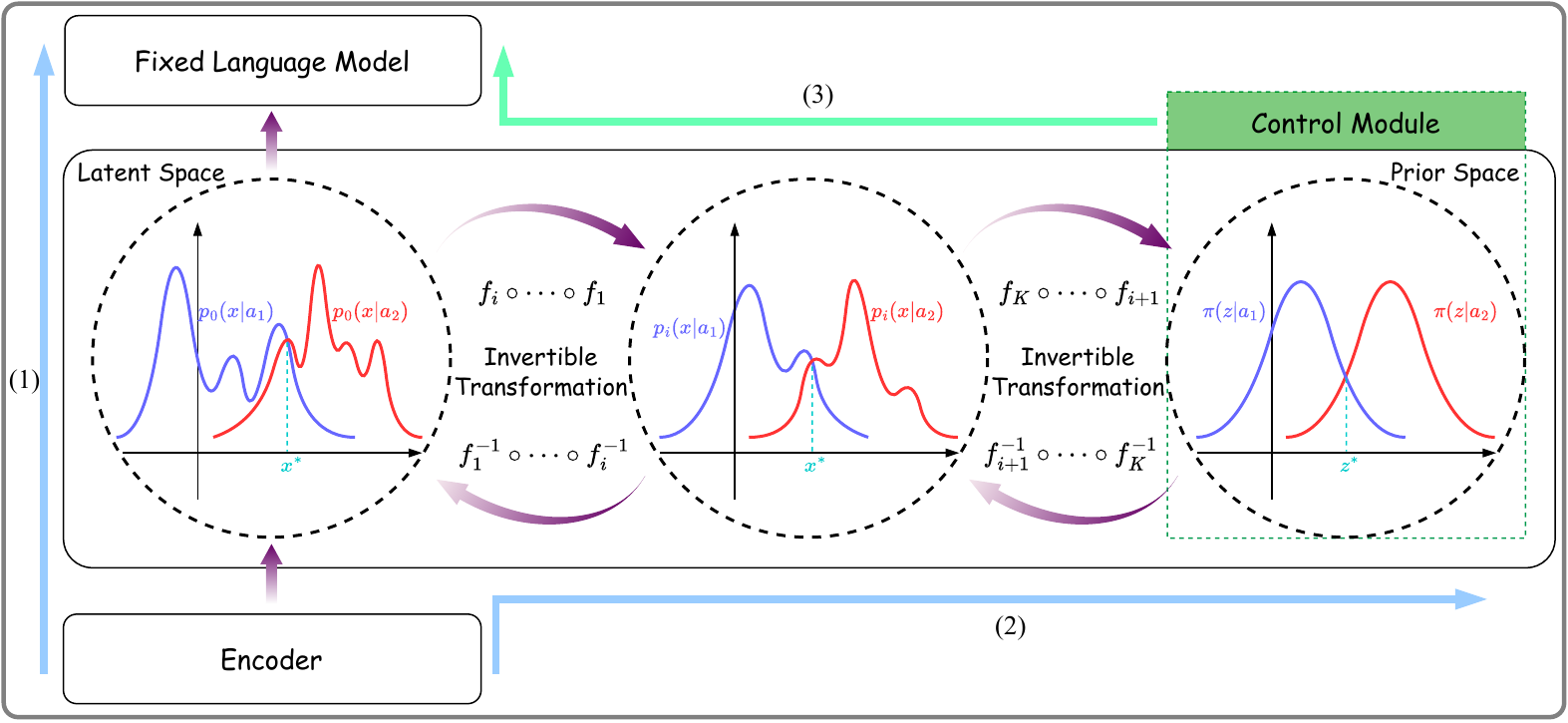}
  }

  \caption{An overview of our framework. Part $(1)$ represents the process of estimating Latent Space, which trains an autoencoder framework by prefix-tuning on the fixed language model. Part $(2)$ denotes the invertible transformation from Latent Space to Prior Space at the training stage. Part $(3)$ consists of two steps: we first operate in Prior Space for control and then feed the effect back into Latent Space to activate the language model.}
  \label{fig:2}

\end{figure*}

\section{Methodology}
As illustrated in Figure \ref{fig:2}, our framework is divided into three parts, where the former two are training phases, and the latter is the generation phase.
\subsection{Estimating the Latent Space}
Given sentence and attribute pairs $\{(s_i, a_i)\}$, we use a learnable encoder to map each sentence to a sample point $x_i \in \mathbb{R}^{n\times1}$, which can activate the fixed language model to reconstruct the same sentence afterward via prefix-tuning. We denote the training loss of this reconstruction target as:

\vspace{-0.2cm}
\begin{equation}
\begin{aligned}
\mathcal{L}_R &=\!-\!\sum\nolimits_i \log p_{\text{LM}}(s_i|\text{Prefix}_i)\\
\text{Prefix}_i &= \text{MLP}_\phi(x_i)\\
x_i&=\text{Encode}_\phi(s_i),
\end{aligned}
\end{equation}
where we can regard each point $x_i$ as being sampled from a continuous Latent Space.
It's worth noting that estimating the Latent Space can be a pre-processing phase that is compatible with any pre-trained auto-encoding structure.

\subsection{Invertible Transformation}
\label{NF}
Normalizing Flow, denoting as $z = f_K \circ \cdots \circ f_1(x) = \mathcal{F}_{\theta}(x)$, maps a point $x_i$ in a complex distribution to the one $z_i\in \mathbb{R}^{n\times1}$ in a simple distribution, such as the Gaussian distribution, with a series of invertible transformations $\{f_i(\cdot)\}$. The probability density function $p(x)$ can be derived as $p(x) =\pi(z)\left|\text{det}\frac{\mathrm{d}\mathcal{F}_\theta(x)}{\mathrm{d}x}\right|$ and the corresponding training target is: $\mathcal{L} = \!-\!\sum_x \log p(x) = \!-\!\sum_x \left[\log \pi(\mathcal{F}_\theta(x)) + \log \left|\text{det}\frac{\mathrm{d}\mathcal{F}_\theta(x)}{\mathrm{d}x}\right|\right]$. See \S \ref{app:normalizingflow} for details about Normalizing Flows.

For controllable text generation, we have to model the conditional probability $p(x|a)$. Therefore, we can decompose the probability as:

\vspace{-0.2cm}
\begin{equation}
\begin{aligned}
p(x) &= \sum\nolimits_{a} p(x|a)p(a)\\
\pi(z) &= \sum\nolimits_{a} \pi(z|a)p(a),
\end{aligned}
\end{equation}
where
$\sum\limits_{a} p(x|a)p(a)\!=\!\sum\limits_{a}\pi(z|a)p(a) \left|\text{det}\frac{\mathrm{d}\mathcal{F}_\theta(x)}{\mathrm{d}x}\right|$.
This means distributions $p(x|a)$ in Latent Space are mapped to the distributions $\pi(z|a)$ in Prior Space through the same invertible transformation $\mathcal{F}_\theta(x)$. 
When each sentence possesses labels of all attributes, which is an ideal supervised situation, we can obtain attribute distributions $p(a)$ and their correlations. However, we usually encounter a semi-supervised situation where a sentence belonging to multiple attributes only has a single attribute label. As a result, we bypass the modeling of $p
(a)$ and set a stricter transformation constraint that $p(x|a) = \pi(z|a) \left|\text{det}\frac{\mathrm{d}\mathcal{F}_\theta(x)}{\mathrm{d}x}\right|$. 
Our target is $\mathcal{L} \!=\!-\!\sum\limits_{(x,a)} \log p(x|a)$, which equals to:

\vspace{-0.2cm}
\begin{equation}
\begin{aligned}
\mathcal{L}\!=\!-\!\sum\limits_{(x,a)}\big[ \log \pi(\mathcal{F}_\theta(x)|a)+\log \left|\text{det}\frac{\mathrm{d}\mathcal{F}_\theta(x)}{\mathrm{d}x}\right|\big].
\end{aligned}
\end{equation}
In this case, we train each attribute independently under the same spatial mapping, where attribute correlations in Latent Space can still be revealed by operation in Prior Space.
It's worth noting that the amount of training data for different attributes should be consistent as possible to ensure the balance of the transformation.
Besides, for the convenience of control, we set covariance matrices $\mathbf{\Sigma}\in \mathbb{R}^{n\times n}$ of prior distributions as diagonal matrices $\sigma^2 = \sigma \sigma^T \mathbf{I}$, where $\pi(z|a) = \mathcal{N}(\mu_a, \sigma_a^2)$.

\subsection{Control in the Prior Space}
In this part, we first prove three significant properties that bridge the Prior and Latent Spaces. Then we introduce how to conduct flexible control in the Prior Space. The three properties ensure that control effect can be fed back into the Latent Space.
\subsubsection{Theoretical Support for Control}
\label{section_theory}
It's worth noting that although the Prior Space is connected to Latent Space with sample-level invertible transformation, the relationship between distributions in the two spaces has not been revealed. Next, we provide three important properties to ensure the effectiveness of controls across the space.

\paragraph{Attribute Preservation} \textit{We define} $z$ \textit{possesses the attribute} $a$ \textit{as in the support set of} $\pi(z|a)$, noted as $z\in \text{supp}(\pi_a)$. \textit{The support of} $\pi_a$ \text{is} $\text{supp}(\pi_a) \!=\! \left\{\pi(z|a)>0, \forall z\right\}$. \textit{Therefore, we have}:

\vspace{-0.4cm}
\begin{equation}
\begin{aligned}
&\exists x,   x = \mathcal{F}_{\theta}^{-1}(z), z \in \text{supp}(\pi_a)\\
&\Rightarrow  p(x|a) = \pi(z|a)\left|\text{det}\frac{\mathrm{d} \mathcal{F}_\theta^{-1}( z)}{\mathrm{d}z}\right|^{-1} > 0\\
&\Rightarrow x \in \text{supp}(p_a),
\end{aligned}
\end{equation}
which means that sampling from $\pi(z|a)$ in Prior Space is equivalent to sampling from $p(x|a)$ in Latent Space, which ensures the effectiveness of single-attribute control in Prior Space.

\paragraph{Intersection Invertibility} 
\textit{The intersection area of multiple attributes} ${a_1, \cdots, a_d}, d\!\leq\!n+1$, \textit{can be defined as the overlapping of their probability density functions} $\{z|\min\left\{\pi(z|a_1),\!\cdots\!,\pi(z|a_d)\right\}\!>\!0\}$.
\textit{In addition, the point where attributes are most tightly combined is considered center of the intersection}: $z^*\!=\!\text{argmax}_z \min\{\pi(z|a_1),\!\cdots\!,\pi(z|a_d)\}$. 
Though there does not necessarily exist a mapping from $z^*$ to the intersection center in Latent Space, we can restrict the region of this mapping to an upper bound.
\textit{Since} $z^*$ \textit{lies in the} $n\text{-}d\text{+}1$ \textit{dimensional subspace} $\mathcal{I}\!=\!\{z|\pi(z|a_1)\!=\!\cdots\!=\!\pi(z|a_d)\}$, \textit{named as Intersection Subspace}, \textit{we can have}:

\vspace{-0.4cm}
\begin{equation}
\begin{aligned}
\forall \hat{z} &\in \mathcal{I}, \exists \hat{x} = \mathcal{F}_\theta^{-1}(\hat{z}),\\ 
p(\hat{x}|a_i) &= \pi(\hat{z}|a_i) \left|\text{det}\frac{\mathrm{d} \mathcal{F}_\theta^{-1}( \hat{z})}{\mathrm{d}\hat{z}}\right|^{-1}\\
 &= \pi(\hat{z}|a_j) \left|\text{det}\frac{\mathrm{d} \mathcal{F}_\theta^{-1}( \hat{z})}{\mathrm{d}\hat{z}}\right|^{-1}\\
 &= p(\hat{x}|a_j), 1\!\leq\!i\!\leq\!d, 1\!\leq\!j\!\leq\!d,
\end{aligned}
\end{equation}
which means that the intersection subspace, where attributes combine most tightly, in Prior Space corresponds to the subspace in Latent Space via bijection, making multi-attribute control effective.

\begin{figure}[t]
\centering
\vspace{-0.5cm}
\includegraphics[width=\columnwidth]{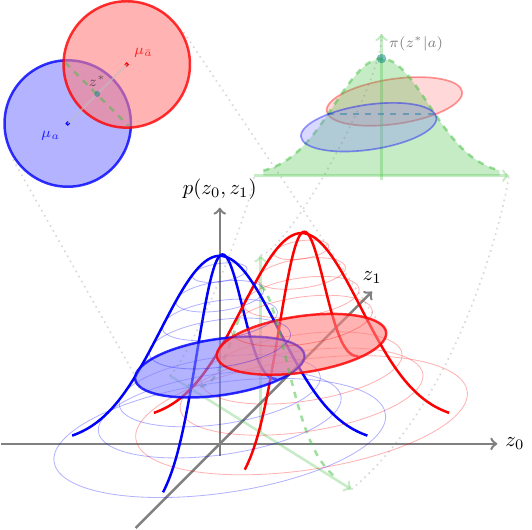}
\vspace{-0.5cm}
\caption{Intersection of two distributions. In the case of isotropy, the intersection of two Gaussian distributions is also a Gaussian distribution, where interpolation of two distribution centers can lie at the intersection center.}
\label{fig:curve}
\end{figure}

\paragraph{Inequality Maintenance} \textit{We define the discrepancy between two attributes concerning the control strength as} $d(x|a_1,\!a_2) = p(x|a_1) - p(x|a_2)$, \textit{measuring the degree of their mutual exclusion. Thus:}

\vspace{-0.5cm}
\begin{equation}
\begin{aligned}
&d(x|a_1,\!a_2) = p(x|a_1) - p(x|a_2)\\
&\ \ =(\pi(z|a_1) - \pi(z|a_2))\left|\text{det}\frac{\mathrm{d} \mathcal{F}_\theta^{-1}( z)}{\mathrm{d}z}\right|^{-1}\\
&\ \ =d(z|a_1,\!a_2)\left|\text{det}\frac{\mathrm{d} \mathcal{F}_\theta^{-1}( z)}{\mathrm{d}z}\right|^{-1}\\
\Rightarrow&\left\{
\begin{aligned}
&\forall z, d(z|a_1,\!a_2)\!>\!0 \Rightarrow d(x|a_1,\!a_2)\!>\!0\\
&\forall z, d(z|a_1,\!a_2)\!<\!0 \Rightarrow d(x|a_1,\!a_2)\!<\!0
\end{aligned}
\right.,
\end{aligned}
\end{equation}
which means inequality of two attributes in Prior Space is also true in Latent Space. Besides, Intersection Subspace of attributes will divide the overlapping of their support sets into two parts, where points in the same part have the same inequality. This is the support for our flexible control strategy.
\subsubsection{Details for Control}
\label{sec:control}

\paragraph{Single-Attribute Control}
Given the \textbf{Attribute Preservation} property, sampling a point $x_a$ related to attribute $a$ in the Latent Space is equivalent to first sampling in Prior Space $z_a \sim \mathcal{N}(\mu_a, \sigma_a^2)$ and then transforming as $x_a = \mathcal{F}_\theta^{-1}(z_a)$. We convert the sampling strategy to $z_a\!=\!\mu_a\!+\!\sigma_a \epsilon, \epsilon\sim \mathcal{N}(\mathbf{0}, \lambda^2 \mathbf{I})$, where $\lambda$ is a hyperparameter\footnote{We will discuss how $\lambda$ influences control strength in \S \ref{exp:1}.}.
\paragraph{Control Strength Adjustment}
Given two mutually exclusive attributes, such as positive $a$ and negative $\bar{a}$ sentiment, sampling an $\alpha$-weighted interpolated point $\tilde{z}$ in Prior Space is $\tilde{z} =\alpha z_{a} + \bar{\alpha} z_{\bar{a}}$, where $\alpha + \bar{\alpha} = \mathbf{1}$. This linear combination is:

\vspace{-0.4cm}
\begin{equation}
\begin{aligned}
\tilde{z} &= (\alpha\mu_a + \bar{\alpha}\mu_{\bar{a}}) + (\alpha\sigma_a\epsilon_a + \bar{\alpha}\sigma_{\bar{a}}\epsilon_{\bar{a}})\\
&= (\alpha\mu_{a} + \bar{\alpha}\mu_{\bar{a}}) + \sqrt{(\alpha\sigma_{a})^2+(\bar{\alpha}\sigma_{\bar{a}})^2} \cdot \epsilon,
\end{aligned}
\end{equation}
which is $\tilde{z}\!\sim\!\mathcal{N}((\alpha\mu_a\!+\!\bar{\alpha}\mu_{\bar{a}}), (\alpha\sigma_{a})^2\!+\!(\bar{\alpha}\sigma_{\bar{a}})^2 \mathbf{I})$. 
As illustrated in the upper left of Figure \ref{fig:curve}, interpolation between $\mu_a$ and $\mu_{\bar{a}}$ is a line 
in Prior Space that passes through the Intersection Subspace\footnote{See \S \ref{sec:appendix1} for the calculation of $\hat{z}$.}, where the intersection point is $\hat{z}\!=\!\alpha^*\mu_a\!+\!\bar{\alpha}^*\mu_{\bar{a}}$. Therefore, sampling with $\hat{z}$ as the center has a great opportunity to sample from the Intersection Subspace in Prior Space, approximate to sampling from the Intersection Subspace in Latent Space based on \textbf{Intersection Invertibility}. 
It is worth noting that when distributions are isotropic, there is $\hat{z}\!=\!z^*$ as in Figure \ref{fig:curve}, which improves the effect of interpolation.
The \textbf{Inequality Maintenance} further ensures that $\alpha\!>\!\alpha^* \!\iff\! p(\mathcal{F}_\theta^{-1}(\tilde{z})|a) \!>\! p(\mathcal{F}_\theta^{-1}(\tilde{z})|\bar{a})$, which means that positive sentiment is guaranteed to be more powerful than negative in Latent Space as long as our weight is larger than $\alpha^*$. 
Our experiment in \S\ref{exp:1} demonstrates that the control strength can be monotonic at a coarse granularity.
When trading off control strength between two polarities, $\alpha$ is usually ranging from $0$ to $1$.
Besides, we can extend the control strength by increasing $\alpha$ to slightly larger than $1$, which equals staying away from attribute $\bar{a}$, as long as it can be guaranteed that points sampled are still within their distribution.

\paragraph{Multi-Attribute Control}
Due to the spatial symmetry of Gaussian distributions, our trained distributions are approximately isotropic when we constrain the covariance matrices to diagonal matrices. This means we can simply deploy the interpolation of each attribute's distributional center as:

\vspace{-0.4cm}
\begin{equation}
\begin{aligned}
z_{i} &= \mu_{i} + \sigma_{i} \epsilon_{i}, \sum\nolimits_i \alpha_i=1, 1\!\leq\!i\!\leq\!d\\
\tilde{z} 
&= \mathcal{N}((\sum\nolimits_i(\alpha_i\mu_i) , \sum\nolimits_i(\alpha_i\sigma_{i})^2 \mathbf{I})
\end{aligned}
\end{equation}

Besides, our Prior Space is compatible with optimization methods. Our optimization process is constrained and the target can be defined as:

\vspace{-0.2cm}
\begin{equation}
\begin{aligned}
\max&\left(\sum\nolimits_i\alpha_i\log\pi(z|a_i)\right)\\
\mathrm{ s.t. } \quad & \forall i\!\neq\!j, \pi(z|a_i)\!=\!\pi(z|a_j),
\end{aligned}
\end{equation}
which is approaching the intersection center $z^*$ in the Intersection Subspace $\mathcal{I}$. We use Lagrange multipliers to handle constraints and sampling with ordinary differential equations as \citet{liu2022composable}.

\vspace{-0.5cm}
\begin{equation}
\begin{aligned}
\mathrm{d}z &= \frac{1}{2} \beta(t)\bigg[ \sum\nolimits_i\alpha_i\nabla_z\log\pi(z|a_i) -\\ &\sum\limits_{i\neq j} \delta_{ij}\nabla_z\big(\log\pi(z|a_i) - \log\pi(z|a_j)\big) \bigg]\mathrm{d}t\\
\delta_{ij} &= \left\{
    \begin{aligned}
    \Omega,\; &\log\pi(z|a_i) - \log\pi(z|a_j) > \tau\\
    \omega,\; &\log\pi(z|a_i) - \log\pi(z|a_j) \leq \tau,
    \end{aligned}
\right. 
\end{aligned}
\end{equation}
where $\Omega \!>\!>\! \omega$ are two hyperparameters and $\tau\!>\!0$ is a threshold. If $\left|\log\pi(z|a_i)\!-\!\log\pi(z|a_j)\right|\!\leq\!\tau$,  $\left|\delta_{ij}\!-\!\delta_{ji}\right|\!=\!0$. If $\left|\log\pi(z|a_i)\!-\! \log\pi(z|a_j)\right|\!>\!\tau$, $\left|\delta_{ij}\!-\!\delta_{ji}\right|\!=\!\Omega\!-\!\omega$. $\beta(t)\!=\!\beta_0\!+\!(\beta_T\!-\!\beta_0)t/T$ is a linear time-variant coefficient, where time $t$ flows forward from $0$ to $T$ and $\mathrm{d}t$ is an infinitesimal positive time step. We provide details about the isotropy of Prior Space and optimization in \S \ref{app:space}.

\section{Experimtents}
\begin{table*}
    \small
    \centering
    \begin{tabular}{l|c|cc|c|cccc|c|c|c}
          \hline 
          
          \hline
          \multirow{2}{*}{\textbf{Methods}} & \multicolumn{3}{c|}{\textbf{Sentiment}↑ (\%)} &\multicolumn{5}{c|}{\textbf{Topic}↑ (\%)} & \textbf{Detox.}↑ & \multirow{2}{*}{\textbf{PPL.}↓} & \multirow{2}{*}{\textbf{Dist.-1/2/3}↑}\\
          \cline{2-9}
           & \textbf{Avg.} &\textbf{Neg.}& \textbf{Pos.}& \textbf{Avg.} & \textbf{W.}&\textbf{S.}& \textbf{B.}&\textbf{T.}& (\%) &  & \\
          \hline
          \hline
          \multicolumn{6}{l}{\quad \textit{Biasing during Decoding}}\\
          \hline
          \textbf{PPLM} & 80.0 & 97.2& 62.7 & 70.6 & 74.9  & 46.5 & 62.4 & 98.6 & 93.2 & 63.2 & 31.1 / 70.9 / 85.9\\
          \hline
          \textbf{GeDi} & 82.3 & 93.9& 70.7 & 83.2& 73.4  & 85.7 & 75.7 & 98.0 & 94.9 & 81.6& 38.1 / 74.0 / 78.4 \\
          \textbf{GeDi}  \footnotesize{\textit{raw}} & 88.4 & 96.6 & 80.2 & 90.8 & 84.3 & 92.6 & 87.1 & 99.2 & 95.4 & 134.1 & 47.5 / 88.9 / 93.0 \\          
          \hline
          \hline
          \multicolumn{6}{l}{\quad\textit{Optimization in the Language Space}}\\
          \hline
          \textbf{MUCOCO} & 75.4 & 95.5 & 55.3 & 73.5 & 56.9 & 67.3 & 72.3 & 97.5 & 94.8 & 381.7 & 22.5 / 49.9 / 64.3 \\
          \hline
          \textbf{Mix\&Match} & 82.8 & 99.2 & 63.3 & 75.6 & 79.5  & 57.4 & 69.6 & \underline{99.3} & \textbf{96.9} & 65.2 & 31.5 / 74.8 / 88.8 \\
          \hline
          \hline
          \multicolumn{6}{l}{\quad\textit{Optimization in the Latent Space}}\\
          \hline
          \textbf{Prefix} & 81.6 & 86.8 & 76.4 & 82.4 & 72.2 & 81.1 & 84.9 & 91.5 & 88.3 & 20.8 & 16.3 / 43.8 / 67.5\\
          \hline
          \textbf{Con. Prefix} & 89.5 & 88.4 & 90.6 & 86.7 & 74.5 & 85.3 & \underline{93.5} & 93.6 & 93.8 & 37.7 & 17.3 / 47.0 / 71.1\\
          \hline
          \textbf{LatentOps}& 91.1 & 88.3 & 93.9 & 69.4 & 54.3 & 61.1 & 72.4 & 89.6 & 94.6 & 58.8 & 13.5 / 48.3 / 62.8\\
          \hline
          \textbf{Discrete} & 92.5 & 99.1 & 85.9 & 90.4 & 84.5 & 95.0 & 84.6 & 97.5 & 90.1 & 46.2 & 36.9 / 76.3 / 87.0\\
          \hline
          \textbf{PriorControl} & \underline{97.1} & \underline{99.9} & \underline{94.3} & \underline{95.9} & \underline{95.5} & \underline{99.3} & 90.2 & 98.7 & 90.7 & 54.3& 29.1 / 70.1 / 86.9\\
          \qquad + extend & \textbf{99.7} & \textbf{99.9} & \textbf{99.5} & \textbf{97.8} & \textbf{97.9} & \textbf{99.4} & \textbf{94.0} & \textbf{99.8} & \underline{95.7} & 54.6 & 29.8 / 70.5 / 86.8\\

          \hline

          \hline
    \end{tabular}

\caption{Automatic Results on Single-Attribute Control. We control on Sentiment (\textbf{Neg}ative and \textbf{Pos}itive), Topic (\textbf{W}orld, \textbf{S}ports, \textbf{B}usiness, and Science/\textbf{T}echnology), and \textbf{Detox}ification independently.}
\label{tab:1}
\end{table*}

\subsection{Tasks and Baselines}
\paragraph{Tasks}
All our experimental setups, including datasets, evaluation metrics, and generation configurations, follow Discrete \cite{gu2022distributional} for fair comparisons. There are IMDb movie reviews \cite{maas-etal-2011-learning}, AGNews dataset \cite{NIPS2015_250cf8b5}, and Jigsaw Toxic Comment Classification Challenge Dataset\footnote{\url{https://www.kaggle.com/c/jigsaw-toxic-comment-classification-challenge/}} for $2$ sentiments, $4$ topics, and $1$ detoxification, respectively. It's worth noting that Discrete randomly samples 10k sentences from each dataset, constituting a minor subset, to balance the data scale for the latent space construction. We directly use this latent space to make a fair comparison.
To evaluate the attribute relevance, we use classifiers trained by Discrete for sentiment and topic, and we utilize the Google Perspective API\footnote{\url{https://www.perspectiveapi.com}} for detoxification. We also measure text quality with Perplexity and Distinctness\cite{li-etal-2016-diversity}.
For human evaluation, each sentence is rated by three professional evaluators for attribute relevance and text fluency. Evaluators rate each item on a scale of 1 to 5, with 5 representing text highly related to the desired attribute or very fluent.
There are $35$ prompts used for text generation, as in PPLM \cite{Dathathri2020Plug}. For single-attribute control, models will generate $5$ completions for each attribute and each prompt, which are $35 \times (2 + 4 + 1) \times 5 = 1225$ sentences. For multi-attribute control, each model generates $35 \times (2 \times 4 \times 1) \times 5 = 1400$ sentences.

\paragraph{Baselines}
\begin{enumerate*}[label=(\Roman*)]
\item \textbf{Biasing during Decoding}:
\textbf{PPLM} \cite{Dathathri2020Plug} accumulates gradients from classifiers as bias signals to influence the language model. \textbf{GeDi} \cite{krause-etal-2021-gedi-generative} biases the decoding process with small conditional generative models.
\item \textbf{Optimization in Language Space}:
\textbf{MUCOCO} \cite{kumar2021controlled} converts the decoding process to multi-objective optimization in language space.
\textbf{Mix\&Match} \cite{mireshghallah-etal-2022-mix} discretely optimizes the sentence in language space by token-level masking and resampling.
\item \textbf{Optimization in Latent Space}:
\textbf{Prefix} \cite{liu2021pre} is the original Prefix-Tuning method which activates the language model to generate attribute-relevant sentences with tunable prefixes.
\textbf{Contrastive Prefix} \cite{qian-etal-2022-controllable} enhances the prefixes through contrastive learning. \textbf{LatentOPs} \cite{liu2022composable} optimizes in latent space with classifiers. 
\textbf{Discrete} \cite{gu2022distributional} uses discrete samples to represent the distribution of attributes in latent space and controls the generation by sampling in relevant areas\footnote{We provide an extra comparison with ChatGPT in \S \ref{sec:chatgpt}.}.

\end{enumerate*}

\subsection{Single-Attribute Control}
We demonstrate the automatic evaluation results on single-attribute control in Table \ref{tab:1}. In addition to the degree of each independent attribute relevance, we compute their average for Sentiment and Topic. Models are grouped with their types of approaches.

\begin{table}[t]
  \small
  \setlength\tabcolsep{2.7pt}
  \centering
      \begin{tabular}{l|cccc|c}
          \hline 
          
          \hline
          \textbf{Methods} & \textbf{Avg.}↑& \textbf{Sent.}↑ & \textbf{Topic}↑ & \textbf{Detox.}↑ &\textbf{Fluency}↑\\
          \hline
          \hline
          \textbf{GeDi \textit{raw}} & 3.28 & 2.66 & 3.40 & 4.08 & 2.81\\
          \hline
          \textbf{Discrete} & 3.42 & 3.28 & 3.42 & 3.68 & 3.47\\
          \hline
          \textbf{PriorControl} & \textbf{4.13} & \textbf{4.05} & \textbf{4.10} & \textbf{4.38} & \textbf{3.61}\\
          \hline

          \hline
      \end{tabular}

\caption{Human Results on Single-Attribute Control.}
  \label{tab:humaneval}

\end{table}

We mainly compare the control methods in the latent space, and the other two technical routes serve as supplementary references.
Biasing methods can achieve decent control at the cost of some fluency. The diversity of their generated sentences is almost the same as the language model, owing to their plug-and-play property during decoding. Besides, we illustrate the raw GeDi without retraining, which is trained on the superset of our dataset. Results show that its performance is affected by the amount of data to some extent.
Optimization methods in language space, elegant in theory, are often troubled by high dimensionality when implemented. 
Optimization in latent space is a compromise strategy where the space dimension is relatively reduced, making the control process more effective but with lower diversity.

\begin{figure}[t]
\centering
\includegraphics[width=0.9\columnwidth]{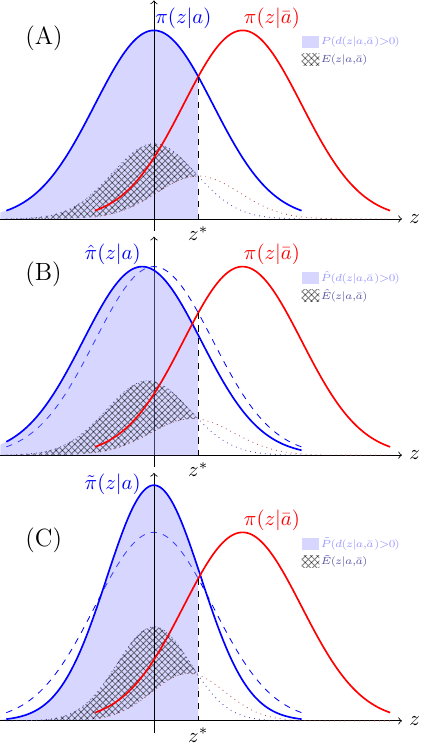}
\caption{The probability density function of two exclusive attributes. (A) represents the original situation. (B) indicates the sampling distribution is slightly away from the undesired one. (C) denotes sampling with a smaller $\lambda$. We measure the control effect by the probability of blue surpassing red $P(d(z|a,\!\bar{a})\!>\!0)$ and the expectation of the difference between blue and red $E(z|a,\!\bar{a})$.}
\label{fig:extend}
\end{figure}


Our method not only enhances the existing latent space optimization method at the level of control strength, with at least 5.0\% and 7.3\% significant improvement over baselines on sentiment and topic. For text quality, our model, sampling points from a Gaussian distribution, can also exceed the original prefix tuning method by 20.5 in the average distinctness.
Our method performs comparable on detoxification because we directly use Discrete's latent space, which is not good at this task.
Compared with Discrete, which assigns the same weight to different sample points, our method can be seen as sampling from the area with higher weights, making our control of higher strength. Although in a continuous space, our sampling will concentrate in a small area with higher probability but similar semantics, making the diversity slightly inferior to Discrete with completely random sampling.
In addition, we show results of human evaluation for single-attribute control in Table \ref{tab:humaneval}, which are almost consistent with automatic evaluation. The agreement of annotators is $0.31$ in Fleiss’ $\kappa$.

Besides, our performance can be further improved by the \textit{extend} control strategy. We can achieve opposite control, as in contrastive learning, by using negative weights when interpolating. Figure \ref{fig:extend}(A) denotes a typical situation where we sample blue points with their probability density function. 
One reason for the suboptimal control effect is that exclusive attributes, denoted as the red distribution, interfere with desired ones, the blue. We can use the probability of blue surpassing red $P(d(z|a,\!\bar{a})\!>\!0)$ and the expectation of the difference between blue and red $E(z|a,\!\bar{a})$ to measure the anti-interference ability in the sampling process\footnote{See \S \ref{app:exclusive} for more details of the two metrics.}.
Figure \ref{fig:extend}(B) shows when our new blue sampling distribution, $\hat{\pi}(z|a)$, is slightly away from the red, surpassing probability and expectation of difference will both increase.
This means the sampling center farther away from interference sources possesses better confidence.
Results of this \textit{extend} control feeding back to the attribute relevance are 2.6, 1.9, and 5.0 improvements on Sentiment, Topic, and Detoxification, respectively.

\subsection{Multi-Attribute Control}

\begin{table*}[ht]
  \small
  \centering
      \begin{tabular}{l|c|ccc|c|c}
          \hline 
          
          \hline
          \textbf{Methods} & \textbf{Average}↑ (\%) & \textbf{Sentiment}↑ (\%) & \textbf{Topic}↑ (\%) & \textbf{Detoxification}↑ (\%) & \textbf{PPL.}↓ &\textbf{Dist.}↑\\
          \hline
          \hline
          \multicolumn{6}{l}{\quad\textit{Biasing during Decoding}}\\
          \hline
          \textbf{PPLM} & 71.0 $\pm$ 21.4 & 64.7 $\pm$ 24.8 & 63.5 $\pm$ 22.7 & 84.9 $\pm$\; 6.5 & 62.6 & 62.0\\
          \hline
          \textbf{GeDi} & 81.4 $\pm$ 14.7 & 76.1 $\pm$ 17.2 & 73.8 $\pm$ 11.3 & 94.2 $\pm$\; 1.9 & 116.6 & 75.1\\
          \hline
          \hline
          \multicolumn{6}{l}{\quad\textit{Optimization in the Language Space}}\\
          \hline
          \textbf{MUCOCO} & 73.9 $\pm$ 24.1 & 65.0 $\pm$ 33.7 & 67.2 $\pm$ 18.3 & 89.5 $\pm$\; 3.5 & 405.6 & 49.7\\
          \hline
          \textbf{Mix\&Match} & 79.7 $\pm$ 21.8 & 73.5 $\pm$ 25.9 & 69.9 $\pm$ 21.1 & \textbf{95.8} $\pm$\; 1.9 & 63.0 & 61.8\\
          \hline
          \hline
          \multicolumn{6}{l}{\quad\textit{Optimization in the Latent Space}}\\
          \hline
          \textbf{Contrastive Prefix} &&&&&\\
          \quad concatenation & 77.2 $\pm$ 18.5 & 67.3 $\pm$ 20.7 & 71.8 $\pm$ 16.5 & 92.6 $\pm$\; 2.9 & 54.6 & 39.9\\
          \quad semi-supervised & 81.3 $\pm$ 16.5 & 74.4 $\pm$ 19.6 & 76.9 $\pm$ 16.7 & 92.7 $\pm$\; 3.5 & 31.9 & 43.3\\
          
          \hline
          \textbf{LatentOps} & 81.6 $\pm$ 15.1 & 82.9 $\pm$ \phantom{0}9.3 & 67.6 $\pm$ 14.7 & 94.2 $\pm$\; 3.5 & 52.2 & 45.4\\
          \hline
          \textbf{Discrete} & 87.4 $\pm$ 10.9 & 86.7 $\pm$ 10.5 & 84.8 $\pm$ 14.2
          & 90.7 $\pm$\; 7.4 & 28.4 & 49.5\\
          \hline
          \textbf{PriorControl} & \underline{89.9} $\pm$ \phantom{0}8.7 & \underline{88.0} $\pm$ 10.6 &  \underline{87.4} $\pm$ \phantom{0}8.5 & 94.3 $\pm$\; 3.2 & 34.7 & 55.5 \\
          \qquad + optim & \textbf{92.2} $\pm$ \phantom{0}8.6 & \textbf{92.5} $\pm$ \phantom{0}8.5 & \textbf{89.3} $\pm$ 11.0 & \underline{94.9} $\pm$\; 3.4 & 29.6 & 51.6\\
          \hline

          \hline
      \end{tabular}
\caption{Automatic Results on Multi-Attribute Control.}
\label{tab:2}
\end{table*}
Automatic evaluation results on multi-attribute control are demonstrated in Table \ref{tab:2}.
We group methods in the same way as single-attribute control, and we add an extra average score for all control combinations. Besides, we demonstrate their standard deviations, which denote the stability of models among different attribute combinations.
Multi-attribute control is more challenging compared to single-attribute control as all models suffer a drop in overall performance.
There are at least 6.3\% and 5.1\% drops in the attribute relevance for Sentiment and Topic. There is little drop in detoxification because this attribute is generally compatible with others.
On one hand, biasing models such as GeDi suffer from a drop not only in control strength but also in the fluency of the generated text, as multiple biasing signals may conflict.
On the other hand, optimization approaches undergo an extra loss in diversity, even including our model, since we have to shrink the variance of the sampling to cut down the decline of the control effect.
As observed in Discrete \cite{gu2022distributional}, this gap between single-attribute control and multi-attribute control is reasonable because different attributes usually combine at sparse edges of their distributions. 
It can also be observed in our mapped prior space that the probability density of the attribute combination region is relatively small.
Compared with Discrete, in addition to control strength, our model possesses better stability according to lower standard deviations. Besides, we outperform the Discrete in diversity because they can only obtain a small number of points in intersection regions, while we can sample from a continuous area.

\section{Analysis}

\subsection{Influence of $\lambda$}
\label{exp:1}

During the sampling stage $\epsilon \sim \mathcal{N}(\mathbf{0}, \lambda^2 \mathbf{I})$, we often anticipate that the obtained points have a higher probability density, which is influenced by $\lambda$.
As mentioned in Figure \ref{fig:extend}, exclusive attributes can interfere with the control effect, and decreasing $\lambda$ is another optional strategy to reduce the interference. We plot the probability density function for $\lambda\!=\!0.8$ in Figure \ref{fig:extend}(C). The probability of blue surpassing red and the expectation of their difference are both larger than the original scores.
Table \ref{tab:adaption} shows the results of $\lambda$'s influence fed back into the latent and language space.
Consistent with the situation in the prior space, attribute relevance increases as $\lambda$ decreases. Besides, since smaller $\lambda$ means concentrating on a smaller area with higher probability density, fluency grows while diversity drops. In addition, we analyze the theoretical influence of $\lambda$ via a toy example of interference between two one-dimensional Gaussian distributions. As in Figure \ref{fig:extend} and \ref{appfig:extend}, we let $\pi(z|a)\!=\!\mathcal{N}(0,1)$ and $\pi(z|\bar{a})\!=\!\mathcal{N}(1.5,1)$. As the $\lambda$ gets smaller, we can see that the probability of the desired attribute surpassing the undesired one $P(d(z|a,\!\bar{a})\!>\!0)$ and the expectation of the difference between the two $E(z|a,\!\bar{a})$ increase, which is consistent to the change of attribute relevance. Therefore, narrowing the sampling area (decreasing $\lambda$) in the prior space will theoretically alleviate the interference from undesired attributes, which can also be reflected in language space, enhancing the control effect in generated sentences.

\begin{table}[t]
  \small
  \centering
  \setlength\tabcolsep{4.5pt}
      \begin{tabular}{c|cc|c|c|c}
          \hline 
          
          \hline
          \multicolumn{3}{l}{\textit{Control on Sentiment}}\\
          \hline
          $\lambda$ & \tiny{$\tilde{P}\hspace{-0.2mm}(\hspace{-0.2mm}d\hspace{-0.2mm}(\hspace{-0.2mm}z\hspace{-0.2mm}|\hspace{-0.2mm}a\hspace{-0.2mm},\hspace{-0.2mm}\!\bar{a})\hspace{-0.5mm}\!>\hspace{-0.5mm}\!0\hspace{-0.2mm})$} & \tiny{$\tilde{E}\hspace{-0.2mm}(\hspace{-0.2mm}z\hspace{-0.2mm}|\hspace{-0.2mm}a\hspace{-0.2mm},\hspace{-0.2mm}\!\bar{a}\hspace{-0.2mm})$} & \textbf{Neg.}/\textbf{Pos.} & \textbf{PPL.}&\textbf{Dist.}\\
          \hline
          1.0 & 0.773 & 0.161 & 99.1 / 78.7 & 85.0 & 64.9\\
          0.9 & 0.798 & 0.171 & 99.4 / 83.0 & 74.7 & 64.6\\
          0.8 & 0.826 & 0.181 & 99.4 / 88.5 & 64.9 & 64.2\\
          0.7 & 0.858 & 0.192 & 99.4 / 92.7 & 59.9 & 63.1\\
          0.6 & 0.894 & 0.205 & 99.9 / 94.3 & 53.9 & 62.0\\
          0.5 & 0.933 & 0.218 & 99.9 / 97.4 & 49.5 & 61.3\\
          0.4 & 0.970 & 0.232 & 99.9 / 99.0 & 45.1 & 60.0\\
          0.3 & 0.994 & 0.246 & 99.9 / 99.0 & 40.3 & 58.2\\
          0.2 & 0.999 & 0.259 & 99.9 / 99.0 & 37.1 & 54.9\\
          0.1 & 1.000 & 0.267 & 99.9 / 99.9 & 34.8 & 52.3\\
          0.0 & 1.000 & 0.269 & 99.9 / 99.9 & 34.3 & 49.9\\
          \hline
         
          \hline
      \end{tabular}
\caption{Results on the $\lambda$'s influence.}
  \label{tab:adaption}

\end{table}

\subsection{Control Strength Adjustment}
We directly adjust the control strength with $\alpha$-interpolation over distribution centers under the approximately isotropic situation. 
As illustrated in \S \ref{NF}, the loss function of invertible transformation is the combination of probability density and Jacobian determinant. Although higher probability in latent space will also tend to be mapped to a higher probability in prior space during the training stage, this tendency is not always guaranteed since the Jacobian determinant can compensate for some loss in probability to obtain a better form of the mapped distribution.
Therefore, there is no strict monotonic relationship between the control strength and the parameter $\alpha$.
Fortunately, as shown in Table \ref{tab:adjustment}, we can observe that the influence of $\alpha$ is approximately monotonic at the coarse-grained level\footnote{See more analyses in \S \ref{app:exclusive}, \ref{app:space}, and \ref{app:optim}.}.


\begin{table}[t]
    \small
  \centering
      \begin{tabular}{c|c|c}
          \hline 
          
          \hline
          \textbf{$\alpha$} & \textbf{Neg} $\rightarrow$ \textbf{Pos} (\%) & \textbf{Toxic} $\rightarrow$ \textbf{NonTox} (\%)\\
          \hline
          1.0 & 99.2 / \phantom{0}0.8 & 79.5 / 20.5\\
          0.9 & 97.1 / \phantom{0}2.9 & 69.4 / 30.6\\
          0.8 & 93.5 / \phantom{0}6.5 & 56.2 / 43.8\\
          0.7 & 88.5 / 11.5 & 49.3 / 50.7\\
          0.6 & 77.7 / 22.3 & 44.2 / 55.8\\
          0.5 & 66.4 / 33.6 & 31.7 / 68.3\\
          0.4 & 49.7 / 50.3 & 23.6 / 76.4\\
          0.3 & 38.1 / 61.9 & 22.0 / 78.0\\
          0.2 & 26.5 / 73.5 & 13.1 / 86.9\\
          0.1 & 19.1 / 80.9 & \phantom{0}9.7 / 90.3\\
          0.0 & 13.7 / 86.3 & \phantom{0}7.4 / 92.6\\
          \hline

          \hline
      \end{tabular}
\caption{Control Strength Adjustment ($\lambda\!=\!1$).}
  \label{tab:adjustment}

\end{table}

\section{Conclusion}
In this work, we present a novel control framework by introducing a well-formed prior space converted from latent space via invertible transformation. 
We further provide some theoretical support to ensure that controls in the prior space can be fed back into the latent space. This allows our framework the potential to generalize to similar situations bothered by high-dimensional and complex latent spaces.
Experimental results confirm the superiority of our model on control effectiveness, control flexibility, and generation quality.

\section*{Limitations}
Our method requires balanced data because all attributes share the same Normalizing Flow. This means that when the training data for one attribute is much larger than others, we need additional training steps to make up such a gap to prevent the Jacobian part of the Normalizing Flow from too much in favor of that attribute. 
In addition, although we can achieve good results on the data scale of $2.5$k or $5$k per attribute, our model does not fit well in few-shot scenarios. We can alleviate this problem by obtaining a sufficient amount of single-attribute labeled data from the style transfer tasks.
In our experiments, each attribute is considered equally important, which may be different from the practical situation. Fortunately, our control strategy is flexible and can be customized for different demands.

\section*{Ethics Statement}
We are fully aware of the potential dangers that text generation techniques may present, such as generating fake, toxic, or offensive content. However, controllable text generation technology is a powerful weapon against harmful information hidden in pre-trained language models, where our study includes text detoxification specifically.
We believe it is beneficial to carry forward research on controllable text generation.
\section*{Acknowledgements}
Xiaocheng Feng is the corresponding author of this work. We thank the anonymous reviewers for their insightful comments. This work was supported by the National Key R\&D Program of China via grant 2020AAA0106502, the National Natural Science Foundation of China (NSFC) via grant 62276078, the Key R\&D Program of Heilongjiang via grant 2022ZX01A32, and the International Cooperation Project of PCL, PCL2022D01.

\bibliography{anthology,custom}
\bibliographystyle{acl_natbib}

\clearpage

\appendix

\section{Defects of the Complex Latent Space}
\label{app:ComplexSpace}
\begin{figure}[h]
\centering
\includegraphics[width=\columnwidth]{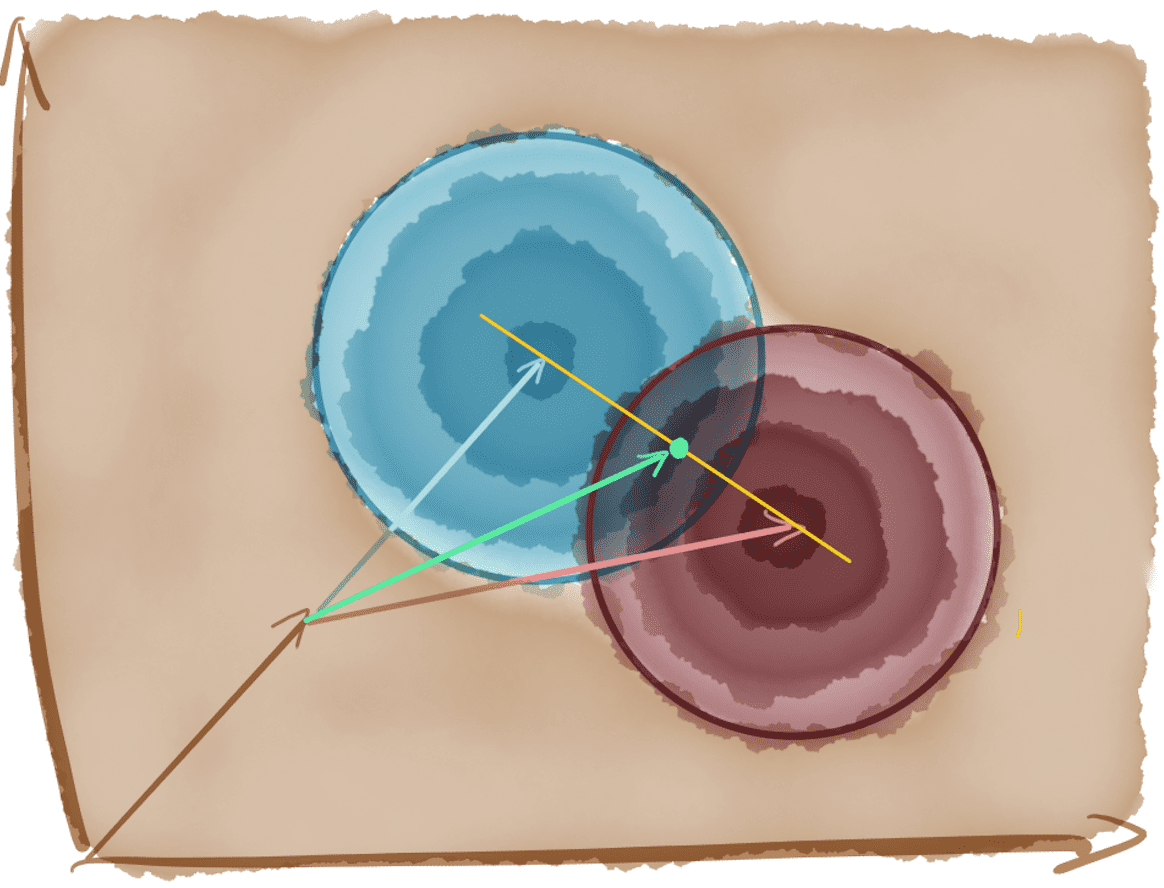}
\includegraphics[width=\columnwidth]{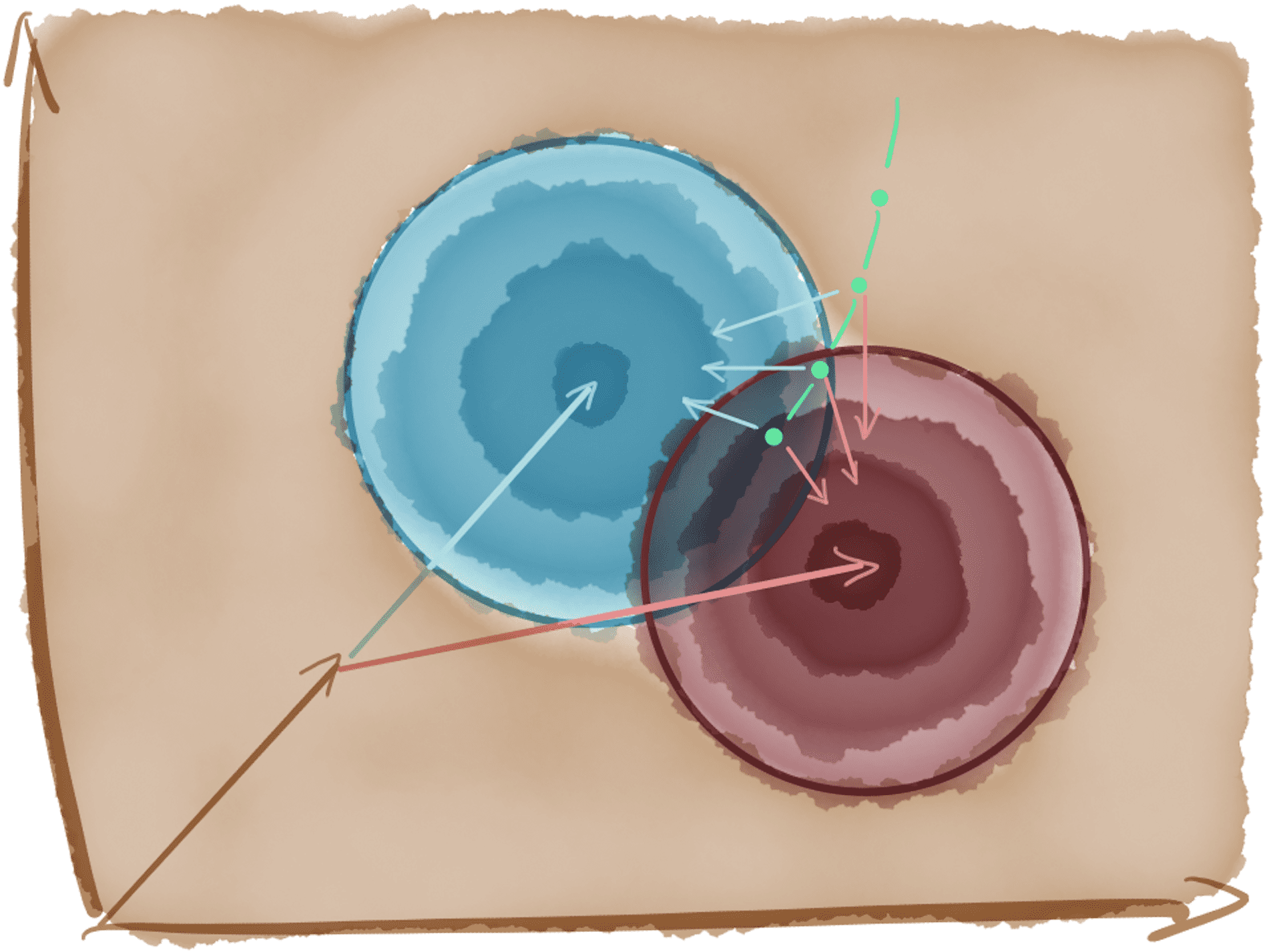}
\caption{Interpolation and Optimization under the ideal oversimplified situation.}
\label{appfig:idealsituation}
\end{figure}

As discussed in Discrete\cite{gu2022distributional}, high-dimensional attribute spaces tend to be asymmetric, anisotropic, and non-convex. Previous work often oversimplifies the spaces of controls into an ideal situation. As demonstrated in Figure \ref{appfig:idealsituation}, interpolation or optimization on two symmetric, isotropic, and convex Gaussian distributions can effortlessly achieve their intersection area.

\begin{figure}[t]
\centering
\includegraphics[width=\columnwidth]{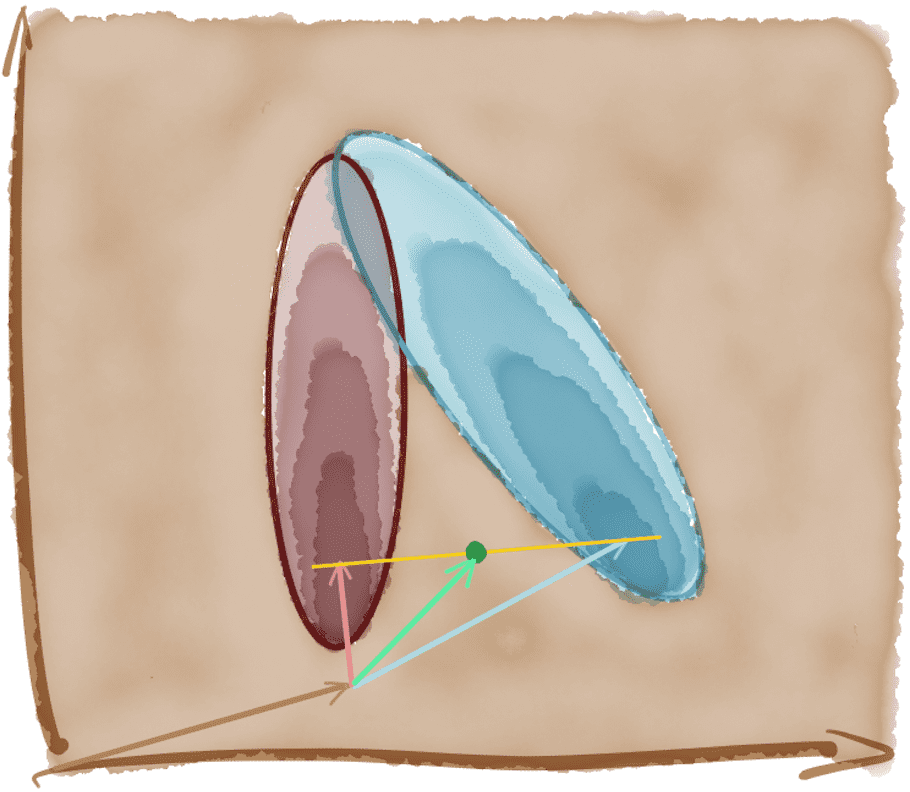}
\includegraphics[width=\columnwidth]{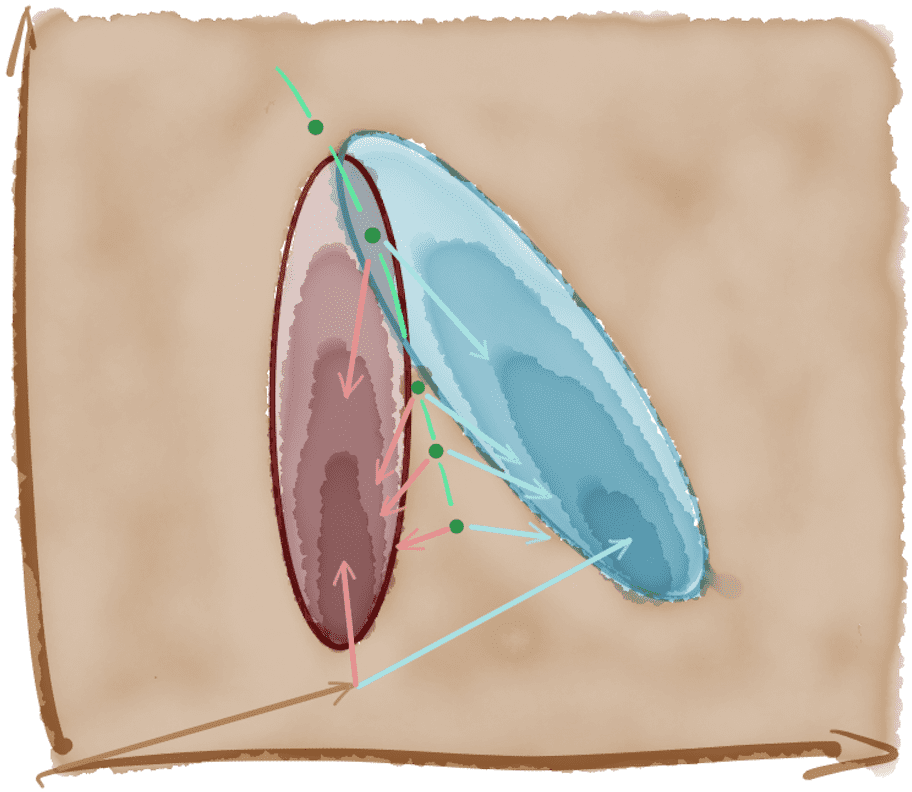}
\caption{Interpolation and Optimization under a more complicated situation.}
\label{appfig:complexsituation}
\end{figure}

\noindent However, it will be completely different when considering a more complicated situation. Figure \ref{appfig:complexsituation} shows the situation when we just make the distributions asymmetric and anisotropic. Both interpolation and optimization will obtain the worst control effect, in which neither distributions are involved nor their intersection. Especially, even if the initialization of the optimization is in the intersection, after a period of iterations, the optimization position will still stop at the saddle point between the distributions, which is outside their support set.
\citet{gu2022distributional} reveal that the Principal Component Analysis (PCA) projections of attribute distributions can sometimes be non-convex. Since the PCA is an operation that preserves convexity, a high-dimensional non-convex distribution may be projected to be convex, but the high-dimensional preimage of a non-convex projection must be non-convex. This means controls in a high-dimensional non-convex space will be even more intractable.

\section{Backgrounds for Normalizing Flows}
\label{app:normalizingflow}
The normalizing Flow, dated back to Non-linear Independent Component Estimation  \cite{dinh2014nice}, is based on the idea that a good representation is one in which the data has an easy-to-model distribution. Since unsupervised learning studies how to capture complex data distributions that have unknown structures, the Normalizing Flow considers a trainable transformation $z=\mathcal{F}_\theta(x)$ of data into a less complicated new space. Following the log-likelihood target in unsupervised learning, this transformation is required to be invertible and the training criterion is derived based on the change of variable rule:
\begin{gather*}
\int p(x)\mathrm{d}x = \int \pi(\mathcal{F}_\theta(x))\mathrm{d}\mathcal{F}_\theta(x)=1
\\
\Longrightarrow p(x) =\pi(\mathcal{F}_\theta(x))\left|\text{det}\frac{\mathrm{d}\mathcal{F}_\theta(x)}{\mathrm{d}x}\right|
\\
\log p(x) = \left[\log \pi(\mathcal{F}_\theta(x)) + \log \left|\text{det}\frac{\mathrm{d}\mathcal{F}_\theta(x)}{\mathrm{d}x}\right|\right],
\end{gather*}
where $\frac{\mathrm{d}\mathcal{F}_\theta(x)}{\mathrm{d}x}$ is the Jacobian matrix of function $\mathcal{F}_\theta$ at $x$. The log-likelihood objective maximizes $p(x)$ for each $x$ in the training data, which approximates simultaneously maximizing the probability density $\pi(\mathcal{F}_\theta(x))$ and the determinant $\left|\text{det}\frac{\mathrm{d}\mathcal{F}_\theta(x)}{\mathrm{d}x}\right|$. 

Since the absolute value of the determinant can be regarded as a scaling factor for the probability density, there is a trade-off between $\pi(\mathcal{F}_\theta(x))$ and $\left|\text{det}\frac{\mathrm{d}\mathcal{F}_\theta(x)}{\mathrm{d}x}\right|$ during training. This means a sample point $x_i$ with a high probability density $p(x_i)$ will not always be mapped to a high probability density $\pi(\mathcal{F}_\theta(x_i))$ because the transformation $\mathcal{F}_\theta(\cdot)$ needs to consider the smoothness of mapping where the determinant $\left|\text{det}\frac{\mathrm{d}\mathcal{F}_\theta(x_i)}{\mathrm{d}x}\right|$ will compensate for this gap of probability density. This situation has little effect when the Normalizing Flow is used as a generative model, however, it is critical to maintain the ratio of the probability density before and after the mapping when performing control. That's why our \S \ref{section_theory} makes sense.

In addition, the key reason for using the Normalizing Flow rather than other generative models is \textbf{invertibility}. Among the current popular generative models, only the Normalizing Flow can achieve the invertible transformation (bijection), which forms the cornerstone of our control framework. For example, the variational autoencoder will construct a fuzzy match between the latent distribution $\pi(z)$ and the sample distribution $p(x)$, and the sample $x$ will randomly correspond to a different $z$ in the same distribution $\pi(z)$ at each training time. This can only be used for single-attribute control, and it will collapse when performing multi-attribute control since we need to decompose and reconstruct attribute distributions from samples that highly rely on bijection. The denoising diffusion probability model shares a similar problem in that a sample $x$ will correspond to an uncertain point $z$ in the latent distribution $\pi(z)$. The generative adversarial network is different in that a latent point $z$ can connect to a determined sample $x$ via the generator. However, the connection can not be reversed, making our control unattainable.

\section{Calculation of $\hat{z}$}
\label{sec:appendix1}

Interpolation of two distribution centers is a line (one-dimensional subspace) where the probability density functions in this subspace are still two Gaussian distributions. That is:
\textit{Our target goes back to solve the equation} $\pi(\hat{z}|a) = \pi(\hat{z}|\bar{a})$ \textit{under one-dimensional situation}.

\begin{figure}[t]
\centering
\includegraphics[width=\columnwidth]{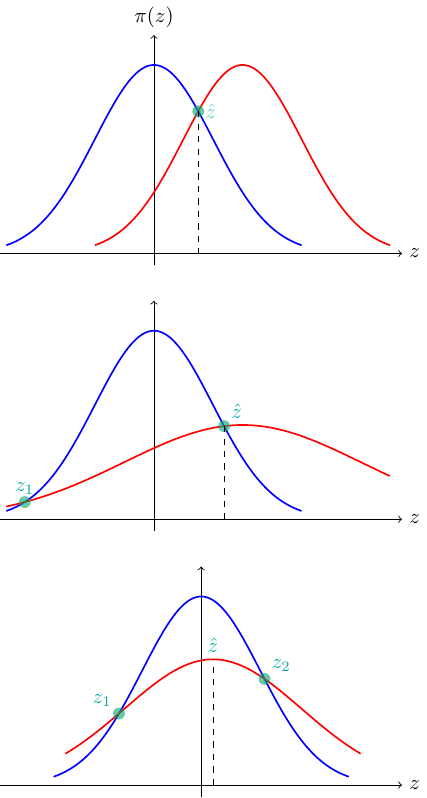}
\caption{Different cases of intersection in one-dimensional subspace.}
\label{fig:inters}
\end{figure}

\vspace{-0.5cm}
$$
\begin{aligned}
&\textit{Given: }\pi(\hat{z}|a) = \pi(\hat{z}|\bar{a})\\
\Rightarrow\ &\mathcal{N}(\hat{z};\mu_a,\sigma^2_a) = \mathcal{N}(\hat{z};\mu_{\bar{a}},\sigma_{\bar{a}}^2)\\
\Rightarrow\ & \log \frac{\exp(-\frac{(\hat{z}-\mu_a)^2}{2\sigma_a^2})}{\sqrt{2\pi}\sigma_a} = \log \frac{\exp(-\frac{(\hat{z}-\mu_{\bar{a}})^2}{2\sigma_{\bar{a}}^2})}{\sqrt{2\pi}\sigma_{\bar{a}}}\\
\Rightarrow\ & \frac{1}{2}\left[\log(\frac{\sigma^2_{\bar{a}}}{\sigma^2_{a}}) - \frac{(\hat{z}-\mu_a)^2}{\sigma^2_a} + \frac{(\hat{z}-\mu_{\bar{a}})^2}{\sigma^2_{\bar{a}}} \right] = 0\\
\Rightarrow\ & \left\{
\begin{aligned}
&A(\hat{z})^2 + B\hat{z} + C = 0\\
&A = -\frac{1}{\sigma_a^2} + \frac{1}{\sigma_{\bar{a}}^2}\\
&B = 2(\frac{\mu_a}{\sigma_a^2} - \frac{\mu_{\bar{a}}}{\sigma_{\bar{a}}^2})\\
&C = \log (\frac{\sigma_{\bar{a}}^2}{\sigma_a^2}) - \frac{\mu_a^2}{\sigma_a^2} + \frac{\mu_{\bar{a}}^2}{\sigma_{\bar{a}}^2}
\end{aligned}
\right.\\
\Rightarrow\ & \begin{aligned}
\Delta &= B^2-4AC\\
&= \frac{4}{\sigma_a^2\sigma_{\bar{a}}^2}\left[(\mu_a-\mu_{\bar{a}})^2 + (\sigma_a^2-\sigma_{\bar{a}}^2)\log(\frac{\sigma_a^2}{\sigma_{\bar{a}}^2})\right]\\
&\geq 0
\end{aligned}\\
\Rightarrow\ & \left\{
\begin{aligned}
&\textit{if } \sigma_a = \sigma_{\bar{a}}, \hat{z} = -\frac{C}{B} = \frac{\mu_a+\mu_{\bar{a}}}{2}\\
&\textit{if } \sigma_a \neq \sigma_{\bar{a}}, \hat{z} = \frac{-B \pm \sqrt{\Delta}}{2A}
\end{aligned}
\right.
\end{aligned}
$$
\textit{According to the derivation above and Figure} \ref{fig:inters}, \textit{when} $\sigma_a = \sigma_{\bar{a}}$, \textit{the} $\hat{z}$ \textit{is simply the midpoint of} $\mu_a$ \textit{and} $\mu_{\bar{a}}$. \textit{When} $\sigma_a \neq \sigma_{\bar{a}}$, \textit{there are usually two solutions for} $\hat{z}$, \textit{and the one we expect needs to be in the interval} $\min(\mu_a,\mu_{\bar{a}})$ \textit{to} $\max(\mu_a,\mu_{\bar{a}})$. \textit{It is worth noting that there may be cases where solutions of} $\hat{z}$ \textit{are both outside this interval, which is caused by the distance between} $\mu_a$ \textit{and} $\mu_{\bar{a}}$ \textit{being too small. In this case, the interval of the two solutions of} $\hat{z}$ \textit{becomes the region where two attributes intersect.}

As illustrated in Figure \ref{fig:inters}, it is complicated to accurately calculate the point where two attributes intersect, even in a one-dimensional case. Fortunately, we can observe that $\hat{z}$ is always between $\mu_a$ and $\mu_{\bar{a}}$. This means we can find an approximate intersection point by adjusting the interpolation parameter in practical use.

\section{Measuring Exclusive Attributes}
\label{app:exclusive}
\begin{figure}[h]
\centering
\includegraphics[width=0.9\columnwidth]{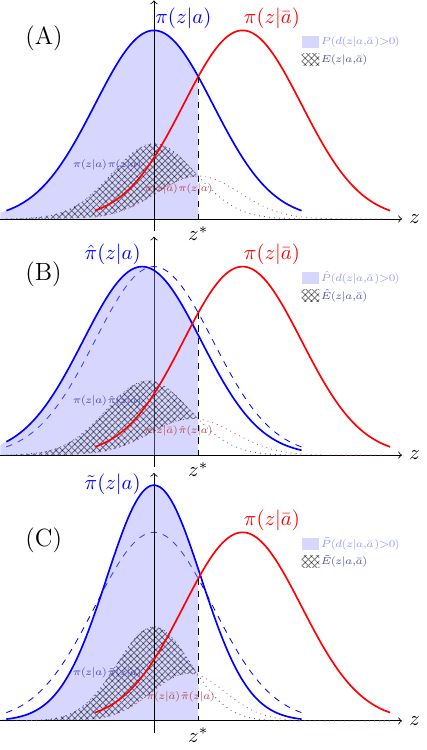}
\caption{The probability density function of two exclusive attributes. (A) represents the original situation, where the blue distribution is $\pi(z|a) = \mathcal{N}(0,1)$ and the red is $\pi(z|\bar{a})=\mathcal{N}(1.5,1)$. (B) indicates the sampling distribution is slightly away from the undesired one, where the new sampling distribution is $\hat{\pi}(z|\bar{a})=\mathcal{N}(\text{-}0.2,1)$. (C) denotes sampling with a smaller $\lambda$, $\tilde{\pi}(z|a)\!=\!\mathcal{N}(0, 0.8^2)$.}
\label{appfig:extend}
\end{figure}
As illustrated in Figure \ref{appfig:extend}, given two exclusive attributes $a$ and $\bar{a}$, attribute $\bar{a}$ will interfere with the effectiveness of $a$ attribute's control.
We measure the control effect by the probability of the blue attribute surpassing the red distribution:

\vspace{-0.2cm}
$$\begin{aligned}P(d(z|a,\bar{a})\!>\!0)\!&=\!\int_{\text{-}\infty}^{z^*}\pi(z|a)\mathrm{d}z\\
\hat{P}(d(z|a,\!\bar{a})\!>\!0)\!&=\!\int_{\text{-}\infty}^{z^*}\hat{\pi}(z|a)\mathrm{d}z\\
\tilde{P}(d(z|a,\!\bar{a})\!>\!0)\!&=\!\int_{\text{-}\infty}^{z^*}\tilde{\pi}(z|a)\mathrm{d}z
\end{aligned}$$ and the expectation of the difference between the blue distribution and the red distribution:

\vspace{-0.2cm}
$$\begin{aligned}E(z|a,\!\bar{a})\!&=\!\int_{\text{-}\infty}^{z^*}\pi(z|a)(\pi(z|a)\!-\!\pi(z|\bar{a}))\mathrm{d}z\\
\hat{E}(z|a,\!\bar{a})\!&=\!\int_{\text{-}\infty}^{z^*}\hat{\pi}(z|a)(\pi(z|a)\!-\!\pi(z|\bar{a}))\mathrm{d}z\\
\tilde{E}(z|a,\!\bar{a})\!&=\!\int_{\text{-}\infty}^{z^*}\tilde{\pi}(z|a)(\pi(z|a)\!-\!\pi(z|\bar{a}))\mathrm{d}z.\end{aligned}$$
It's worth noting that due to the symmetry of the Gaussian distribution, when the red distribution is on the left side of the blue, it will only affect the integral's starting point and end point rather than the result.
In Figure \ref{appfig:extend}, part (A) is the original situation where $P(d(z|a,\!\bar{a})\!>\!0)\!\approx\!0.773$ and $E(z|a,\!\bar{a})\!\approx\!0.161$. Part (B) is the extending trick in \S \ref{sec:control} that keeps the sampling distribution slightly away from the exclusive distribution by a distance of $0.2$, where $\hat{P}(d(z|a,\!\bar{a})\!>\!0)\!\approx\!0.829$ and $\hat{E}(z|a,\!\bar{a})\!\approx\!0.171$. Note that the offset needs to be balanced between staying away from interference and maintaining the original sampling area. Part (C) is concentrating the sampling area with a smaller $\lambda$, where $\tilde{P}(d(z|a,\!\bar{a})\!>\!0)\!\approx\!0.826$ and $\tilde{E}(z|a,\!\bar{a})\!\approx\!0.181$. It's apparent that these two sampling strategies are compatible and follow the same principle: reallocate higher weights to more reliable regions which possess higher probability density and less noise.

Next, we analyze the effect of these strategies fed back into the latent space. 
For part (B), we assume the offset is $s\!<\!0$, which means $\hat{\pi}(z|a)\!=\!\pi(z\!-\!s|a)$. Therefore, we can prove that $\hat{P}(d(z|a,\!\bar{a})\!>\!0)\!>\!P(d(z|a,\!\bar{a})\!>\!0)$:

\vspace{-0.5cm}
$$
\begin{aligned}
&\hat{P}(d(z|a,\!\bar{a})\!>\!0)=\int_{\text{-}\infty}^{z^*}\hat{\pi}(z|a)\mathrm{d}z\\
\qquad &=\int_{\text{-}\infty}^{z^*}\pi(z\!-\!s|a)\mathrm{d}z =\int_{\text{-}\infty}^{z^*\!-\!s}\pi(z|a)\mathrm{d}z\\
\qquad &=\int_{\text{-}\infty}^{z^*}\pi(z|a)\mathrm{d}z + \int_{z^*}^{z^*\!-\!s}\pi(z|a)\mathrm{d}z\\
\qquad &>\int_{\text{-}\infty}^{z^*}\pi(z|a)\mathrm{d}z.
\end{aligned}
$$
Based on the \textbf{Inequality Maintenance}, there exists no $x\!<\!x^*$ that can make $\mathcal{F}_{\theta}(x)\!>\!\mathcal{F}_{\theta}(x^*)$. This means points in the interval $(\text{-}\infty, x^*]$ from the latent space have a one-to-one correspondence with points in $(\text{-}\infty, z^*]$ from the prior space. Therefore, when mapping to latent space, we have $\int_{\text{-}\infty}^{x^*}p(x|a)\mathrm{d}x\!=\!\int_{\text{-}\infty}^{z^*}\pi(z|a)\frac{\mathrm{d}\mathcal{F}_{\theta}(x)}{\mathrm{d}x}\mathrm{d}x\!=\!\int_{\text{-}\infty}^{z^*}\pi(z|a)\mathrm{d}z$. Thus, we have $\int_{\text{-}\infty}^{z^*}\hat{\pi}(z|a)\mathrm{d}z\!=\!\int_{\text{-}\infty}^{x^*}p(x|a)\mathrm{d}x\!+\!\int_{z^*}^{z^*\!-\!s}\pi(z|a)\mathrm{d}z\!>\!\int_{\text{-}\infty}^{x^*}p(x|a)\mathrm{d}x$. 
It's worth noting that interval $[z^*, z^*\!-\!s]$ is not guaranteed to correspond to interval $[x^*, \mathcal{F}_{\theta}^{\text{-}1}\left(z^*\!-\!s\right)]$. 
As a result, points sampled from $\hat{\pi}(z|a)$ possess higher probability density in the latent space.
For part (C), which is similar, we have $\tilde{P}(d(z|a,\!\bar{a})\!>\!0)\!>\!P(d(z|a,\!\bar{a})\!>\!0)$:

\vspace{-0.5cm}
$$
\begin{aligned}
&\tilde{P}(d(z|a,\!\bar{a})\!>\!0)=\int_{\text{-}\infty}^{z^*}\tilde{\pi}(z|a)\mathrm{d}z\\
\qquad &=\int_{\text{-}\infty}^{z^*}\frac{\pi(\frac{z}{\lambda}|a)}{\lambda}\mathrm{d}z =\int_{\text{-}\infty}^{\frac{z^*}{\lambda}}\pi(z|a)\mathrm{d}z\\
\qquad &=\int_{\text{-}\infty}^{z^*}\pi(z|a)\mathrm{d}z\!+\!\int_{z^*}^{\frac{z^*}{\lambda}}\pi(z|a)\mathrm{d}z\\
\qquad &>\int_{\text{-}\infty}^{z^*}\pi(z|a)\mathrm{d}z.
\end{aligned}
$$
Therefore, we have $\int_{\text{-}\infty}^{z^*}\tilde{\pi}(z|a)\mathrm{d}z\!=\!\int_{\text{-}\infty}^{x^*}p(x|a)\mathrm{d}x\!+\!\int_{z^*}^{\frac{z^*}{\lambda}}\pi(z|a)\mathrm{d}z\!>\!\int_{\text{-}\infty}^{x^*}p(x|a)\mathrm{d}x$, which means the probability of sampled points in latent space is monotonically increasing as $\lambda$ decreases.
We can also observe the same phenomenon from the perspective of $E(z|a,\!\bar{a})$. However, their proof requires the integration of Gaussian distributions, which is complex that we skip here.

\section{Hyperparameters and Details}
We directly leverage the latent space provided by Discrete \cite{gu2022distributional}, which is implemented on the \textit{Huggingface Transformers} package\footnote{\url{https://github.com/huggingface/transformers}}. The encoder is initialized with Bert-base-uncased, and the fixed decoder uses GPT2-medium. Each training sentence will be tokenized with WordPiece tokenizer from Bert and Byte-Pair Encoding tokenizer from GPT2 before input to encoder and decoder, respectively. As in Discrete, we perform mean pooling on outputs of the encoder and convert them to 768-dimensional latent representations, which are points in the latent space. Afterward, latent representations will be mapped to the prefix with a dimension of $20 \times 24 \times 2 \times 1024$, where $20$ is the prefix sequence length, $24$ is the number of hidden layers in GPT2-medium, $2$ represents one key and one value, and $1024$ is the size of hidden states in GPT2-medium. Our invertible transformation works like a plug-and-play module on the latent space and we implement it with the \textit{FrEIA} package\footnote{\url{https://github.com/vislearn/FrEIA}}. The normalizing flow contains $8$ layers, each of which is composed of two linear layers and one activation layer. Normalization flows preserve the dimensionality of the input vectors, which means that our prior space has the same dimension as the latent space of $768$. In addition, we follow LatentOps \cite{liu2022composable} and utilize the \textit{torchdiffeq} package\footnote{\url{https://github.com/rtqichen/torchdiffeq}} for solving the ordinary differential equations in the prior space.

During the training stage, the parameters of the encoder, decoder, and prefix mapping module are fixed and initialized with ones from Discrete. We only train the parameters of the Normalizing Flow with half-precision mode on one NVIDIA A100 80GB GPU, where the batch size is 100. In our setting, the random seed is $0$, the optimizer is AdamW with a learning rate of 1e-4, all $8$ attributes are trained together in different batches, and the training steps are $300000$ which spends about 9 hours. 

\begin{table}[h]
\small
\centering
\begin{tabular}{l|c|c}
    \hline
    
    \hline
    \textbf{Combination} & \textbf{Weight} $w_i$ & $\lambda$\\
    \hline
    \hline
    Neg. \& World \& NonTox. & $2:12:1$ & 0.5\\
    Neg. \& Sports \& NonTox. & $2:6:1$ & 0.5\\
    Neg. \& Business \& NonTox. & $2:16:1$ & 0.5\\ 
    Neg. \& Sci./Tech. \& NonTox. & $2:1:5$ & 0.5\\
    Pos. \& World \& NonTox. & $14:16:0.2$ & 0.2\\
    Pos. \& Sports \& NonTox. & $28:20:0.2$ & 0.2\\
    Pos. \& Business \& NonTox. & $20:26:0.2$ & 0.1\\
    Pos. \& Sci./Tech. \& NonTox. & $6:1:1$ &0.5\\
    \hline
    
    \hline
\end{tabular}
\caption{Hyperparameters for Multi-Attribute Control.}
\label{apptab:hyper}
\end{table}

During the inference phase for single-attribute control, we choose $\lambda\!=\!0.6$ to balance the control strength, fluency, and diversity. In \textit{extend} mode, our sampling center is slightly away from the adversarial attribute by a distance of 0.2. We can move away from the interpolation center for aspects with more than two attributes, such as the topic aspect. For multi-attribute control, we utilize a specialized list of hyperparameters, weight $w_i$ and $\lambda$, for each combination of attributes in Table \ref{apptab:hyper}, where $\alpha_i\!=\!\frac{w_i}{\sum_j w_j}$.
Our customized hyperparameters are aimed at balancing the control effect among attributes and the trade-off between diversity and attribute relevance.
After mapping samples back to the latent space, the text generation process is the same as Discrete, where the sequence length is set to 50. The entire evaluation process for each attribute combination takes less than 1 minutes, allowing us to fine-grain the search for satisfying hyperparameters, where the maximum trial number for each attribute combination is $10$. For constrained optimization in Intersection Subspace, the hyperparameters are $\Omega\!=\!0.3$, $\omega\!=\!0.01$, $\tau\!=\!\text{8e-5}$, $\beta_0=20$, $\beta_T=0.1$, and $T=1$.

Same as Discrete, 35 prompts we used in the inference stage are following the PPLM setting with 20 from its bag-of-word setting and 15 from its discriminator setting:
\begin{itemize}
\item \textbf{PPLM-Bow}: ``In summary'', ``This essay discusses'', ``Views on'', ``The connection'', ``Foundational to this is'', ``To review,'', ``In brief,'', ``An illustration of'', ``Furthermore,'', ``The central theme'', ``To conclude,'', ``The key aspect'', ``Prior to this'', ``Emphasised are'', ``To summarise'', ``The relationship'', ``More importantly,'', ``It has been shown'', ``The issue focused on'', ``In this essay''.
\item \textbf{PPLM-Discrim}: ``Once upon a time'', ``The book'', ``The chicken'', ``The city'', ``The country'', ``The horse'', ``The lake'', ``The last time'', ``The movie'', ``The painting'', ``The pizza'', ``The potato'', ``The president of the country'', ``The road'', ``The year is 1910.''.
\end{itemize}

Detailed setting of baselines: \begin{enumerate*}[label=(\Roman*)]
\item \textbf{Biasing during Decoding}:
For \textbf{PPLM}, we only retrain its classifier heads on our datasets while keeping all other original settings. For \textbf{GeDi}, we provide two versions. One is retrained on our dataset and another uses the raw parameters since their dataset is the superset of ours.
\item \textbf{Optimization in Language Space}:
\textbf{MUCOCO} provides a solution for custom classification constraints, and thus we train these classifiers on our dataset. \textbf{Mix\&Match} is relatively complex as it can not generate long sentences from scratch with the mask language model Bert. Worse still, as a method based on sampling, it is somewhat dependent on initialization. Therefore, we use sentences generated by PPLM as the starting sentences and let Mix\&Match slowly polish the text by itself in iterations. 
\item \textbf{Optimization in Latent Space}:
We reproduce \textbf{Contrastive Prefix} and achieve comparable results. For \textbf{LatentOps}, we retrain both their VAE structure and the classifiers for optimization. We directly use \textbf{Discrete} as we follow their settings.
\end{enumerate*}
For a fair comparison, we unify the pre-trained language model to GPT2-medium (345M parameters) except for Mix\&Match using Bert-large (340M parameters). 

\newpage
\section{Statistics of Prior Space}
\label{app:space}
\subsection{Isotropic and Anisotropic}
\label{sec:appendix2}

We analyze $\sigma$ of different attribute's Gaussian distribution $\mathcal{N}(\mu_a, \sigma_a^2)$ in Table \ref{apptab:1}. We demonstrate the maximum, minimum, average, and standard deviation values among all dimensions for each $\sigma$.
The maximum differences of $\sigma$s are around $0.1$, and the standard deviations are all less than $0.02$, in which case we consider the distributions to be approximately isotropic.
\begin{table}[t]
    \centering
    \small
    \begin{tabular}{l|c|c|c|c}
        \hline
        
        \hline
        \multirow{2}{*}{\textbf{Attribute}} & \multicolumn{4}{c}{$\sigma_i$} \\
        \cline{2-5}
         & $\max$ & $\min$ & $\mathop{\text{avg}}$ & $\text{std}$\\
        \hline
        \hline
        Negative    & 0.886 & 0.756 & 0.800 & 0.018\\
        Positive    & 0.889 & 0.760 & 0.801 & 0.018\\
        World       & 0.848 & 0.737 & 0.782 & 0.018\\
        Sports      & 0.837 & 0.728 & 0.776 & 0.018\\
        Business    & 0.851 & 0.738 & 0.783 & 0.018\\
        Sci./Tech.  & 0.853 & 0.737 & 0.784 & 0.018\\
        Toxic       & 0.853 & 0.740 & 0.783 & 0.017\\
        NonTox.     & 0.853 & 0.747 & 0.790 & 0.017\\
        \hline
        
        \hline
    \end{tabular}
    \caption{$\sigma$ of Different Attributes.}
    \label{apptab:1}
\end{table}

\begin{figure}[t]
\centering
\includegraphics[width=\columnwidth]{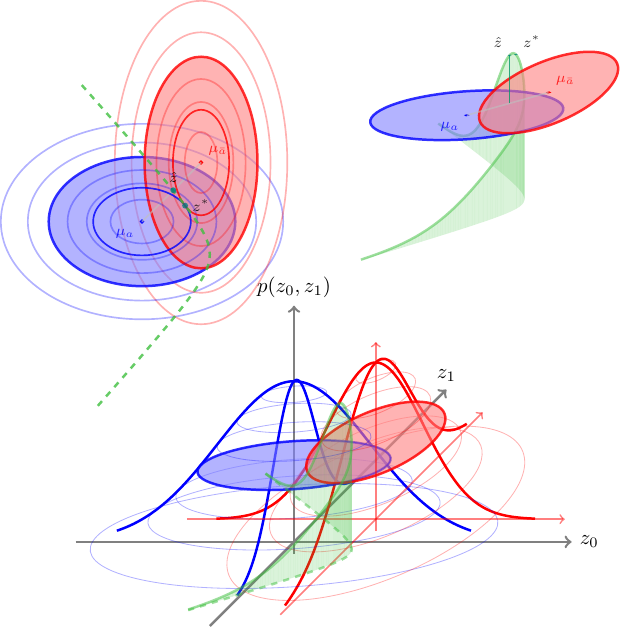}
\caption{Intersection of two anisotropic distributions.}
\label{fig:anisotropic}
\end{figure}

Furthermore, we plot the situation in Figure \ref{fig:anisotropic} when two 2-dimensional anisotropic distributions intersect. We set up an example: $\mu_a\!=\!(0,0)^T,\sigma_a\!=\!(1.3,0.9)^T,\mu_{\bar{a}}\!=\!(1,1)^T,\sigma_{\bar{a}}\!=\!(0.8,1.5)^T$. At this time, the intersection subspace is still a one-dimensional subspace, but it becomes a hyperbola rather than a straight line. Besides, the interpolation method can only obtain a suboptimal intersection point $\hat{z}$, where the optimal point lies in $z^*$. As shown in \S \ref{sec:control}, optimization methods are expected to achieve $z^*$ with $\hat{z}$ as the initialization, higher probability density as the target, and Intersection Subspace as constraints.

Next, we provide the derivation for the intersection of two distributions in 2-dimensional space.

\vspace{-0.5cm}
$$
\begin{aligned}
&\textit{Given: }\pi(\hat{z}|a) = \pi(\hat{z}|\bar{a})\\
\Rightarrow\ &\mathcal{N}(\hat{z};\mu_a,\sigma^2_a) = \mathcal{N}(\hat{z};\mu_{\bar{a}},\sigma_{\bar{a}}^2)\\
\Rightarrow\ & \log \frac{\exp(-\frac{(\hat{z}_1-\mu_{a,1})^2}{2\sigma_{a,1}^2}-\frac{(\hat{z}_2-\mu_{a,2})^2}{2\sigma_{a,2}^2})}{2\pi\cdot\sigma_{a,1}\cdot\sigma_{a,2}}\\
= & \log \frac{\exp(-\frac{(\hat{z}_1-\mu_{\bar{a},1})^2}{2\sigma_{\bar{a},1}^2}-\frac{(\hat{z}_2-\mu_{\bar{a},2})^2}{2\sigma_{\bar{a},2}^2})}{2\pi\cdot\sigma_{\bar{a},1}\cdot\sigma_{\bar{a},2}}\\
\Rightarrow\ & \log(\frac{\sigma^2_{a,1}\!\cdot\!\sigma^2_{a,2}}{\sigma^2_{\bar{a},1}\!\cdot\!\sigma^2_{\bar{a},1}}) + \left[\frac{(\hat{z}_1-\mu_{a,1})^2}{\sigma^2_{a,1}} - \frac{(\hat{z}_1-\mu_{\bar{a},1})^2}{\sigma^2_{\bar{a},1}}\right] \\
& + \left[\frac{(\hat{z}_2-\mu_{a,2})^2}{\sigma^2_{a,2}} - \frac{(\hat{z}_2-\mu_{\bar{a},2})^2}{\sigma^2_{\bar{a},2}}\right] = 0\\
\Rightarrow\ & \log(\frac{\sigma^2_{a,1}\!\cdot\!\sigma^2_{a,2}}{\sigma^2_{\bar{a},1}\!\cdot\!\sigma^2_{\bar{a},1}}) + \frac{A_1\hat{z}_1^2\!+\!B_1\hat{z}_1+C_1}{\sigma^2_{a,1}\cdot\sigma^2_{\bar{a},1}} \\
& + \frac{A_2\hat{z}_2^2\!+\!B_2\hat{z}_2+C_2}{\sigma^2_{a,2}\cdot\sigma^2_{\bar{a},2}} = 0,\\
& \left\{
\begin{aligned}
&A_i = \sigma^2_{\bar{a},i}-\sigma^2_{a,i}\\
&B_i = 2(\mu_{\bar{a},i}\cdot\sigma^2_{a,i}-\mu_{a,i}\cdot\sigma^2_{\bar{a},i})\\
&C_i = \mu^2_{a,i}\cdot\sigma^2_{\bar{a},i}-\mu^2_{\bar{a},i}\cdot\sigma^2_{a,i}
\end{aligned}
\right.\\
\Rightarrow\ & \frac{\left(\hat{z}_1+\frac{B_1}{2A_1}\right)^2}{\sigma^2_{a,1}\!\cdot\!\sigma^2_{\bar{a},1}/A1} + \frac{\left(\hat{z}_2+\frac{B_2}{2A_2}\right)^2}{\sigma^2_{a,2}\!\cdot\!\sigma^2_{\bar{a},2}/A2} = K,\\
& K = \log(\frac{\sigma^2_{\bar{a},1}\!\cdot\!\sigma^2_{\bar{a},1}}{\sigma^2_{a,1}\!\cdot\!\sigma^2_{a,2}}) + \frac{B_1^2-4A_1C_1}{4A_1\cdot\sigma^2_{a,1}\!\cdot\!\sigma^2_{\bar{a},1}} \\
& + \frac{B_2^2-4A_2C_2}{4A_2\cdot\sigma^2_{a,2}\!\cdot\!\sigma^2_{\bar{a},2}}\\
\Rightarrow\ & \frac{\left(\hat{z}_1-M_1\right)^2}{P_1} + \frac{\left(\hat{z}_2-M_2\right)^2}{P_2} = 1,\\
& \left\{
\begin{aligned}
&M_i = -\frac{B_i}{2A_i}\\
&P_i = \frac{\sigma^2_{a,i}\cdot\sigma^2_{\bar{a},i}}{A_i\cdot K}
\end{aligned}
\right.
\end{aligned}
$$
\begin{table*}[ht]
  \small
  \centering
      \begin{tabular}{l|c|ccc|c|c}
          \hline 
          
          \hline
          \textbf{Methods} & \textbf{Average}↑ (\%) & \textbf{Sentiment}↑ (\%) & \textbf{Topic}↑ (\%) & \textbf{Detoxification}↑ (\%) & \textbf{PPL.}↓ &\textbf{Dist.}↑\\
          \hline
          \hline
          \textbf{Discrete} & 87.4 $\pm$ 10.9 & 86.7 $\pm$ 10.5 & 84.8 $\pm$ 14.2
          & 90.7 $\pm$\; 7.4 & 28.4 & 49.5\\
          \hline
          \textbf{PriorControl} & 89.9 $\pm$ \phantom{0}8.7 & 88.0 $\pm$ 10.6 &  87.4 $\pm$ \phantom{0}8.5 & 94.3 $\pm$\; 3.2 & 34.7 & 55.5 \\
          \quad + uncons optim & \underline{91.8} $\pm$ \phantom{0}9.7& \underline{89.7} $\pm$ 11.9 & \textbf{90.1} $\pm$ 10.4 & \textbf{95.5} $\pm$\; 3.0 & 29.9 & 52.1\\
          \quad + optim & \textbf{92.2} $\pm$ \phantom{0}8.6 & \textbf{92.5} $\pm$ \phantom{0}8.5 & \underline{89.3} $\pm$ 11.0 & \underline{94.9} $\pm$\; 3.4 & 29.6 & 51.6\\
          \hline

          \hline
      \end{tabular}
\caption{Automatic Results on Multi-Attribute Control for Different Optimization Strategies.}
\label{tab:anisotropic}
\end{table*}
When $A_1=0$ and $A_2=0$, the Intersection Subspace is a straight line as in the isotropic situation. When $A_1=0, A_2\neq0$ or $A_1\neq0, A_2=0$, the Intersection Subspace becomes a parabola. When $A_1\neq0, A_2\neq0$ and $A_1\times A_2>0$, the Intersection Subspace is an ellipse.
As in Figure \ref{fig:anisotropic}, when $A_1\neq0, A_2\neq0$ and $A_1\times A_2 < 0$, it becomes a hyperbola. When multiple distributions intersect in a high-dimensional space, the formula of Intersection Subspace can be generalized according to the conic section above.  

\subsection{Optimize in Prior Space}
We demonstrate in Table \ref{tab:anisotropic} the effect of optimizing in the prior space. \textit{Unconstrained Optimization} represents a simplified target which drops the constraints as: $\mathrm{d}z\!=\!\frac{1}{2} \beta(t)\left[\sum\nolimits_i \alpha_i \nabla_z\log\pi(z|a_i)\right]\mathrm{d}t$. Compared with \textbf{PriorControl}, the improvement brought by optimization mainly comes from two parts: one is the large sampling area of \textbf{PriorControl} leads to some low-probability samples, and the other is the interpolation of \textbf{PriorControl} cannot perfectly achieve the optimal point $z^*$. This means although our model is approximately isotropic, it cannot completely ignore the influence of differences in various dimensions. Furthermore, the marginal improvement from constraints means that optimization does not need to worry about problems such as saddle points, which means the shape of our prior space is satisfyingly simple.

In addition, we illustrate the detailed results of multi-attribute combinations in Table \ref{apptab:detailcombination}. It is rare for the attribute relevance to degrade after optimization when constrained in the intersection subspace. However, without these constraints, the optimization process becomes unstable and more likely to decay. Since these degradations are usually marginal, we consider that dropping constraints can improve optimization speed when the prior space is simple and regular.

\begin{table*}[h]
    \centering
    \small
    \begin{tabular}{l|cc|cccc|c}
          \hline 
          
          \hline
          \multirow{2}{*}{\textbf{Methods}} & \multicolumn{2}{c|}{\textbf{Sentiment} (\%)} &\multicolumn{4}{c|}{\textbf{Topic} (\%)} & \multirow{2}{*}{\textbf{Detox.} (\%)}\\
           & \textbf{Neg.}& \textbf{Pos.}& \textbf{World}&\textbf{Sports}& \textbf{Business}&\textbf{Sci./Tech.}&\\
          \hline
          \hline

          \multirow{8}{*}{\textbf{Discrete}}
          & 69.7\phantom{↑} & - & 71.7\phantom{↑} & - & - & - & 84.1\phantom{↑}\\
          & 78.6\phantom{↑} & - & - & 80.0\phantom{↑} & - & - & 80.2\phantom{↑}\\
          & 99.9\phantom{↑} & - & - & - & 96.7\phantom{↑} & - & 96.8\phantom{↑}\\
          & 92.8\phantom{↑} & - & - & - & - & 98.0\phantom{↑} & 81.7\phantom{↑}\\
          & - & 80.5\phantom{↑} & 58.0\phantom{↑} & - & - & - & 95.1\phantom{↑}\\
          & - & 84.7\phantom{↑} & - & 86.6\phantom{↑} & - & - & 94.5\phantom{↑}\\
          & - & 87.6\phantom{↑} & - & - & 91.7\phantom{↑} & - & 98.1\phantom{↑}\\
          & - & 99.7\phantom{↑} & - & - & - & 96.1\phantom{↑} & 95.4\phantom{↑}\\
          \hline
          \hline
          
          \multirow{8}{*}{\textbf{PriorControl}}
          & 94.6\phantom{↑} & - & 90.2\phantom{↑} & - & - & - & 90.1\phantom{↑}\\
          & 96.5\phantom{↑} & - & - & 97.4\phantom{↑} & - & - & 93.0\phantom{↑}\\
          & 91.4\phantom{↑} & - & - & - & 88.5\phantom{↑} & - & 97.6\phantom{↑}\\
          & 99.6\phantom{↑} & - & - & - & - & 97.1\phantom{↑} & 88.8\phantom{↑}\\
          & - & 79.5\phantom{↑} & 80.1\phantom{↑} & - & - & - & 94.8\phantom{↑}\\
          & - & 82.4\phantom{↑} & - & 79.5\phantom{↑} & - & - & 95.7\phantom{↑}\\
          & - & 65.6\phantom{↑} & - & - & 72.5\phantom{↑} & - & 98.2\phantom{↑}\\
          & - & 94.3\phantom{↑} & - & - & - & 93.8\phantom{↑} & 96.2\phantom{↑}\\
          \hline
                    
          \multirow{8}{*}{\tabincell{l}{\textbf{PriorControl}\\ \quad + uncons optim}}
          & 97.7\textcolor{sgreen}{↑} & - & 99.2\textcolor{sgreen}{↑} & - & - & - & 92.4\textcolor{sgreen}{↑}\\
          & 97.9\textcolor{sgreen}{↑} & - & - & 98.5\textcolor{sgreen}{↑} & - & - & 95.1\textcolor{sgreen}{↑}\\
          & 97.9\textcolor{sgreen}{↑} & - & - & - & 96.7\textcolor{sgreen}{↑} & - & 98.6\textcolor{sgreen}{↑}\\
          & 99.9\textcolor{sgreen}{↑} & - & - & - & - & 98.1\textcolor{sgreen}{↑} & 89.2\textcolor{sgreen}{↑}\\
          & - & 83.2\textcolor{sgreen}{↑} & 75.7\textcolor{sred}{↓} & - & - & - & 95.8\textcolor{sgreen}{↑}\\
          & - & 75.6\textcolor{sred}{↓} & - & 83.7\textcolor{sgreen}{↑} & - & - & 97.0\textcolor{sgreen}{↑}\\
          & - & 67.1\textcolor{sgreen}{↑} & - & - & 72.2\textcolor{sred}{↓} & - & 98.2\textcolor{sred}{↓}\\
          & - & 89.7\textcolor{sred}{↓} & - & - & - & 90.1\textcolor{sred}{↓} & 97.3\textcolor{sgreen}{↑}\\
          \hline
          
          \multirow{8}{*}{\tabincell{l}{\textbf{PriorControl}\\ \quad + optim}}
          & 97.9\textcolor{sgreen}{↑} & - & 98.3\textcolor{sgreen}{↑} & - & - & - & 90.5\textcolor{sgreen}{↑}\\
          & 98.4\textcolor{sgreen}{↑} & - & - & 98.5\textcolor{sgreen}{↑} & - & - & 93.4\textcolor{sgreen}{↑}\\
          & 97.3\textcolor{sgreen}{↑} & - & - & - & 96.9\textcolor{sgreen}{↑} & - & 98.5\textcolor{sgreen}{↑}\\
          & 99.9\textcolor{sgreen}{↑} & - & - & - & - & 99.7\textcolor{sgreen}{↑} & 89.1\textcolor{sgreen}{↑}\\
          & - & 89.5\textcolor{sgreen}{↑} & 79.4\textcolor{sred}{↓} & - & - & - & 95.4\textcolor{sgreen}{↑}\\
          & - & 84.5\textcolor{sgreen}{↑} & - & 73.7\textcolor{sred}{↓} & - & - & 96.8\textcolor{sgreen}{↑}\\
          & - & 74.2\textcolor{sgreen}{↑} & - & - & 73.1\textcolor{sgreen}{↑} & - & 98.4\textcolor{sgreen}{↑}\\
          & - & 98.0\textcolor{sgreen}{↑} & - & - & - & 95.2\textcolor{sgreen}{↑} & 97.3\textcolor{sgreen}{↑}\\      
          
          \hline

          \hline

          \hline
    \end{tabular}

\caption{Detailed Combination Results on Multi-Attribute Control. }
  \label{apptab:detailcombination}
\end{table*}

\subsection{Distance between Distributions}
We also analyze the distances between distributions in \Cref{apptab:2,apptab:3,apptab:4}, which are automatically learned without human guidance. The distance is calculated as the absolute difference for each corresponding dimension between two distributions. Table \ref{apptab:2} and Table \ref{apptab:3} show the average and maximum values of the distance in each dimension, respectively.
The large discrepancy between the average and maximum values indicates that the distances in most dimensions are small. And the differences between distributions are mainly determined by the few dimensions with the largest distances. Therefore, we additionally show the average value of the top-5 dimensions in Table \ref{apptab:4}. Consistent with the intuition, we can observe that the distance between two mutually exclusive attributes is relatively large, such as negative-positive sentiments and toxic-nontoxic. Furthermore, topics are generally farther from the positive sentiment than the negative one, which is in line with our experimental results in Table \ref{apptab:detailcombination}. The business topic is a counterexample that performs better on control strength with negative sentiment than positive while its distribution stays closer to the positive one. Compared to the performance in \textbf{Discrete}, we assume that this may be due to our probability density estimation on the business topic not being very good.
\begin{table}[t]
    \small
    \centering
    \setlength\tabcolsep{2pt}
    \begin{tabular}{c|cc|cccc|c|}
        \hline
        
        \hline
        \multirow{2}{*}{\textbf{Average}} & \multicolumn{2}{c|}{\textbf{Sentiment}} &\multicolumn{4}{c|}{\textbf{Topic}} & \textbf{Detox.} \\
        \cline{2-8}
        &\textbf{Neg.}& \textbf{Pos.}& \textbf{W.}&\textbf{S.}& \textbf{B.}&\textbf{T.}& \textbf{Tox.}\\
        \hline
        \hline
        Positive    & 0.199 & - & & & & &\\
        World       & 0.135 & 0.156 & - & & & &\\
        Sports      & 0.166 & 0.203 & 0.184 & - & & &\\
        Business    & 0.176 & 0.131 & 0.176 & 0.224 & - & &\\
        Sci./Tech.  & 0.163 & 0.142 & 0.166 & 0.248 & 0.128 & - &\\
        Toxic       & 0.130 & 0.178 & 0.124 & 0.149 & 0.178 & 0.192 &\\
        NonTox.     & 0.161 & 0.124 & 0.153 & 0.207 & 0.116 & 0.100 & 0.187\\
        \hline
        
        \hline
    \end{tabular}
    \caption{Average of $|\mu_{a_i}\!-\!\mu_{a_j}|$ between Attributes.}
    \label{apptab:2}
\end{table}

\begin{table}[t]
    \small
    \centering
    \setlength\tabcolsep{2pt}
    \begin{tabular}{c|cc|cccc|c|}
        \hline
        
        \hline
        \multirow{2}{*}{\textbf{Max}} & \multicolumn{2}{c|}{\textbf{Sentiment}} &\multicolumn{4}{c|}{\textbf{Topic}} & \textbf{Detox.} \\
        \cline{2-8}
        &\textbf{Neg.}& \textbf{Pos.}& \textbf{W.}&\textbf{S.}& \textbf{B.}&\textbf{T.}& \textbf{Tox.}\\
        \hline
        \hline
        Positive    & 0.776 & - & & & & &\\
        World       & 0.544 & 0.620 & - & & & &\\
        Sports      & 0.571 & 0.735 & 0.651 & - & & &\\
        Business    & 0.751 & 0.452 & 0.798 & 0.794 & - & &\\
        Sci./Tech.  & 0.565 & 0.645 & 0.857 & 0.848 & 0.620 & - &\\
        Tox.        & 0.525 & 0.639 & 0.493 & 0.559 & 0.632 & 0.733 & -\\
        NonTox.     & 0.637 & 0.504 & 0.635 & 0.697 & 0.458 & 0.439 & 0.702\\
        \hline
        
        \hline
    \end{tabular}
    \caption{Maximum of $|\mu_{a_i}\!-\!\mu_{a_j}|$ between Attributes.}
    \label{apptab:3}
\end{table}

\begin{table}[t]
    \small
    \centering
    \setlength\tabcolsep{2pt}
    \begin{tabular}{c|cc|cccc|c|}
        \hline
        
        \hline
        \multirow{2}{*}{\textbf{Top-5}} & \multicolumn{2}{c|}{\textbf{Sentiment}} &\multicolumn{4}{c|}{\textbf{Topic}} & \textbf{Detox.} \\
        \cline{2-8}
        &\textbf{Neg.}& \textbf{Pos.}& \textbf{W.}&\textbf{S.}& \textbf{B.}&\textbf{T.}& \textbf{Tox.}\\
        \hline
        \hline
        Positive    & 0.700 & - & & & & &\\
        World       & 0.529 & 0.572 & - & & & &\\
        Sports      & 0.533 & 0.670 & 0.624 & - & & &\\
        Business    & 0.617 & 0.438 & 0.699 & 0.750 & - & &\\
        Sci./Tech.  & 0.541 & 0.557 & 0.686 & 0.795 & 0.545 & - &\\
        Toxic       & 0.491 & 0.605 & 0.472 & 0.535 & 0.607 & 0.709 & -\\
        NonTox.     & 0.561 & 0.464 & 0.600 & 0.666 & 0.404 & 0.365 & 0.630\\
        \hline
        
        \hline
    \end{tabular}
    \caption{Top-5 of $|\mu_{a_i}\!-\!\mu_{a_j}|$ between Attributes.}
    \label{apptab:4}
\end{table}

\section{Optimize in Different Latent Spaces}
\label{app:optim}

\begin{table*}[h]
    \small
    \centering
    \setlength\tabcolsep{5.5pt}
    \begin{tabular}{l|c|cc|c|cccc|c|c|c}
          \hline 
          
          \hline
          \multirow{2}{*}{\textbf{Methods}} & \multicolumn{3}{c|}{\textbf{Sentiment}↑ (\%)} &\multicolumn{5}{c|}{\textbf{Topic}↑ (\%)} & \textbf{Detox.}↑ & \multirow{2}{*}{\textbf{PPL.}↓} & \multirow{2}{*}{\textbf{Dist.-1/2/3}↑}\\
          \cline{2-9}
           & \textbf{Avg.} &\textbf{Neg.}& \textbf{Pos.}& \textbf{Avg.} & \textbf{W.}&\textbf{S.}& \textbf{B.}&\textbf{T.}& (\%) &  & \\
          \hline
          \hline
          \multicolumn{6}{l}{\quad\textit{Optimization in Simple Latent Space}}\\
          \hline
          \textbf{LatentOps}& 91.1 & 88.3 & 93.9 & 69.4 & 54.3 & 61.1 & 72.4 & 89.6 & 94.6 & 58.8 & 13.5 / 48.3 / 62.8\\
          \hline
          \hline
          \multicolumn{6}{l}{\quad\textit{Optimization in Complex Latent Space}}\\
          \hline
          \hline
          \textbf{Discrete} & 88.2 & 98.5\phantom{↓} & 77.8\phantom{↓} & 89.7 & 84.5\phantom{↓} & 95.0\phantom{↓} & 84.5\phantom{↓} & 94.7\phantom{↓} & 88.7\phantom{↓} & 46.4 & 35.5 / 77.7 / 89.2\\
          \textbf{DiscreteOps} & 89.2 & 97.7\textcolor{sred}{↓} & 80.7\textcolor{sgreen}{↑} & 89.6 & 84.0\textcolor{sred}{↓} & 95.0\textcolor{sred}{↓} & 84.2\textcolor{sred}{↓} & 95.2\textcolor{sgreen}{↑} & 90.3\textcolor{sgreen}{↑} & 47.5 & 35.8 / 79.1 / 90.0\\
          \hline
          \textbf{Discrete} \textit{best} & 92.5 & 99.1\phantom{↓} & 85.9\phantom{↓} & 90.4 & 84.5\phantom{↓} & 95.0\phantom{↓} & 84.6\phantom{↓} & 97.5\phantom{↓} & 90.1\phantom{↓} & 46.2 & 36.9 / 76.3 / 87.0\\
          \hline
          \hline
          \multicolumn{6}{l}{\quad\textit{Optimization in Prior Space}}\\
          \hline
          \textbf{Prior} \scriptsize{($\lambda\!=\!1.0$)} & 88.9 & 99.1\phantom{↓} & 78.7\phantom{↓} & 81.4 & 70.2\phantom{↓} & 92.4\phantom{↓} & 66.9\phantom{↓} & 96.2\phantom{↓} & 91.8\phantom{↓} & 85.0 & 31.6 / 73.9 / 89.2\\
          \textbf{PriorOps} & 87.9 & 97.8\textcolor{sred}{↓} & 77.9\textcolor{sred}{↓} & 84.2 & 74.3\textcolor{sgreen}{↑} & 92.2\textcolor{sred}{↓} & 72.3\textcolor{sgreen}{↑} & 97.9\textcolor{sgreen}{↑} & 91.7\textcolor{sred}{↓} & 85.2 & 31.2 / 73.6 / 89.1\\
          \hline
          \textbf{PriorControl} & 97.1 & 99.9\phantom{↓} & 94.3\phantom{↓} & 95.9 & 95.5\phantom{↓} & 99.3\phantom{↓} & 90.2\phantom{↓} & 98.7\phantom{↓} & 90.7\phantom{↓} & 54.3 & 29.1 / 70.1 / 86.9\\
          \qquad + extend & \underline{99.7} & \underline{99.9}\phantom{↓} & \underline{99.5}\phantom{↓} & \underline{97.8} & \underline{97.9}\phantom{↓} & \underline{99.4}\phantom{↓} & \underline{94.0}\phantom{↓} & \underline{99.8}\phantom{↓} & \textbf{95.7}\phantom{↓} & 54.6 & 29.8 / 70.5 / 86.8\\
          \qquad + optim & \textbf{99.8} & \textbf{99.9}\phantom{↓} & \textbf{99.6}\phantom{↓} & \textbf{99.6} & \textbf{99.9}\phantom{↓} & \textbf{99.7}\phantom{↓} & \textbf{98.9}\phantom{↓} & \textbf{99.9}\phantom{↓} & \underline{94.3}\phantom{↓} & 34.8 & 23.1 / 57.4 / 75.8\\

          \hline

          \hline
    \end{tabular}

\caption{Results on Single-Attribute Control for the Optimization in Different Spaces.}
\label{tab:diff_optim}
\end{table*}

As demonstrated in Table \ref{tab:diff_optim}, we analyze how the optimization method performs in different spaces. \textbf{LatentOps} \cite{liu2022composable} utilizes ordinary differential equations to optimize sampling points in simple latent spaces constructed by the VAE structure. Their latent spaces only require the dataset of the corresponding aspect each time for single-attribute control. Therefore, they perform well in aspects with only two attributes, like sentiment and detoxification, while they are mediocre in complex aspects, such as topic. We migrate the optimization method to the complex latent space of Discrete, named as \textbf{DiscreteOps}. \textbf{Discrete}'s space is specially designed for the combination of multiple attributes, where there exist eight attributes. For single-attribute control, we randomly sample a set of points in the corresponding attribute's training data as prefixes for text generation. We experiment with several random seeds and pick the best one for each attribute as the upper bound, \textbf{Discrete} \textit{best}. Since optimization requires good initialization, as described in \textbf{LatentOps}, we use a random seed with average performances, i.e., \textbf{Discrete}, as \textbf{DiscreteOps}'s initialization. It's interesting to observe that optimization is more likely to improve performance when there is a large gap between \textbf{Discrete} and \textbf{Discrete} \textit{best}. On the contrary, when \textbf{Discrete} is close to the upper bound, optimization may degrade the attribute relevance. We think this is because classifiers are not good tools for probability density modeling, where most region of the space is not in the classifier's domain of definition, making the optimization process coarse. This phenomenon can also be observed after migrating to the prior space. \textbf{PriorControl} we use in the main experiment sets $\lambda\!=\!0.6$, and we let the points sampled when $\lambda\!=\!1.0$ as the initialization of \textbf{PriorOps}. When the energy function composed of the classifier is used as the optimization target in the prior space, the attribute correlation of generated text cannot surpass the performance of \textbf{Discrete}. However, when we keep $\lambda\!=\!1.0$ and replace the optimization target with the Gaussian distribution of corresponding attribute in the prior space, which is \textbf{PriorControl} + \textit{optim}, the control strength can exceed the + \textit{extend} method at the cost of diversity. This reveals that our work provides not only a better conditional probability density estimation method but also a better control framework that is compatible with current control strategies.

\section{Compare with ChatGPT}
\label{sec:chatgpt}
Based on the principle of a fair comparison, we use \textit{gpt2-medium} as the language model, which is consistent with baselines. In this section, we briefly test the controllability of ChatGPT (early version before January 1, 2023), which is the most powerful conditional generative language model at present. The \textit{magic spell} we use to activate ChatGPT is ``\textit{Generate 5 sentences containing 50 words with} [ATTRIBUTE] \textit{and start with} `[PROMPT]'.'' The [ATTRIBUTE] is selected from negative sentiment, positive sentiment, world topic, sports topic, business topic, technology topic, and non-toxicity. The [PROMPT] is from the 35 prompts we used in the experiments. As illustrated in Table \ref{tab:chatgpt}, ChatGPT can accurately identify the task of attribute control and achieve impressive performance, especially on sentiment and detoxification. Because there are a large number of training datasets for both aspects. When facing attributes such as topics, which possess a relatively small amount of data, ChatGPT can only make limited associations based on keywords of the topic while can not control from a more abstract level. It is obvious that text it generated has a strong fluency that almost reaches the human level. However, although deliberately replacing words during decoding, it still lacks diversity in the scenario of large-scale open-ended text generation.
We also show some cases in Table \ref{tab:casecompare_negative} and Table \ref{tab:casecompare_world}. For negative sentiment control, ChatGPT can generate fluent sentences with high attribute strength. Interestingly, it is insensitive to structural controls such as sentence length. For world topic control, ChatGPT tends to associate some keywords from the word \textit{world}, such as \textit{economic}, \textit{community}, and \textit{global}, while cannot generate sentences that feel like something happened somewhere in the world. Compared to our GPT2-based framework, ChatGPT can generate sentences with better quality and fewer factual inconsistency issues.

\begin{table*}[t]
    \small
    \centering
    \begin{tabular}{l|c|cc|c|cccc|c|c|c}
          \hline 
          
          \hline
          \multirow{2}{*}{\textbf{Methods}} & \multicolumn{3}{c|}{\textbf{Sentiment}↑ (\%)} &\multicolumn{5}{c|}{\textbf{Topic}↑ (\%)} & \textbf{Detox.}↑ & \multirow{2}{*}{\textbf{PPL.}↓} & \multirow{2}{*}{\textbf{Dist.-1/2/3}↑}\\
          \cline{2-9}
           & \textbf{Avg.} &\textbf{Neg.}& \textbf{Pos.}& \textbf{Avg.} & \textbf{W.}&\textbf{S.}& \textbf{B.}&\textbf{T.}& (\%) &  & \\
          \hline
          \hline
          
          \textbf{ChatGPT} & \textbf{99.9} & 99.8 & \textbf{99.9} & 77.5 & 60.7 & 72.2 & 78.7 & 98.5 & \textbf{98.4} & 14.3 & 21.4 / 46.8 / 56.9\\

          \hline
          \textbf{PriorControl} & 97.1 & \underline{99.9} & 94.3 & \underline{95.9} & \underline{95.5} & \underline{99.3} & \underline{90.2} & \underline{98.7} & 90.7 & 54.3& 29.1 / 70.1 / 86.9\\
          \qquad + extend & \underline{99.7} & \textbf{99.9} & \underline{99.5} & \textbf{97.8} & \textbf{97.9} & \textbf{99.4} & \textbf{94.0} & \textbf{99.8} & \underline{95.7} & 54.6 & 29.8 / 70.5 / 86.8\\

          \hline

          \hline
    \end{tabular}

\caption{Automatic Results on Single-Attribute Control compared with ChatGPT.}
\label{tab:chatgpt}
\end{table*}

\begin{table*}
    \centering
    \begin{tabular}{|p{7.55cm}||p{7.55cm}|}
        \hline
        
        \hline
        \textbf{ChatGPT} & \textbf{PriorControl} \\
        \hline
        \hline
        \multicolumn{2}{|l|}{Generate 5 sentences containing 50 words with negative sentiment and start with ‘The movie’.}\\
        \hline
        The movie explores the growing sense of despair and hopelessness that is taking hold of our society. & The movie was terrible. It's a shame that such blatant disregard of genre conventions was so prevalent in this movie. The plot was ridiculous, the dialogue was cliched, and the sound quality was poor.\\
        \specialrule{0em}{2pt}{2pt}
        The movie discusses the increasing levels of stress and anxiety that are affecting our mental health and well-being. & The movie is one of the worst I have ever seen. The plot is a bit like that of a scam involving some of the female characters. There should be no such film for this week.\\
        \specialrule{0em}{2pt}{2pt}
        The movie examines the growing sense of disillusionment and frustration with our political system and its inability to address the pressing issues of the day. & The movie is so bad, I could hardly believe what I saw. The plot is ridiculous, the direction is terrible and there are no lessons in this movie.\\
        \specialrule{0em}{2pt}{2pt}
        The movie looks at the mounting evidence that our society is becoming increasingly divided and unequal, with negative consequences for all of us. & The movie was horrific. Even though most people are just trying to avoid leaving home in the Gaza Strip, they have no reason to dislike him. The scenes were so scripted and acted like a joke.\\
        \specialrule{0em}{2pt}{2pt}
        The movie explores the pervasive sense of discontent and dissatisfaction with the state of our society, and the urgent need for action to address these issues. & The movie is so bad that it almost got caught on TV. The characters are terrible, the plot contrived, and utterly hopeless to watch. There is no connection between the stars, the actual history of this movie, or the countless pointless deaths.\\

        \hline
        
        \hline
    \end{tabular}
    \caption{ChatGPT's cases for negative sentiment control.}
    \label{tab:casecompare_negative}
\end{table*}

\begin{table*}
    \centering
    \begin{tabular}{|p{7.55cm}||p{7.55cm}|}
        \hline
        
        \hline
        \textbf{ChatGPT} & \textbf{PriorControl} \\
        \hline
        \hline
        \multicolumn{2}{|l|}{Generate 5 sentences containing 50 words with world topic and start with ‘The city’.}\\
        \hline
        The city, as a hub of economic activity, cultural exchange, and technological innovation, is discussed as a potential model for global development and progress. & The city of Abu Dhabi has declared independence from Iraq, amid a furious Islamist rally on Monday and an increasingly provocative UN official. "It is such vile political propaganda," said the Islamic cleric-turned politician quoted by Reuters yesterday.\\
        \specialrule{0em}{2pt}{2pt}
        The city, with its diverse and often marginalized communities, is examined as a potential site for the protection and promotion of human rights. & The city of Afghanistan completed a test for its constitution, according to an official statement. Related videos: A suicide bomber killed a soldier in the Iraq war.\\
        \specialrule{0em}{2pt}{2pt}
        The city, with its complex and interconnected systems, is considered as a potential laboratory for addressing global challenges such as climate change and pandemics. & The city of Cheuba entered a marathon fundraising campaign Saturday, amid an international outcry. Provided by Reuters.\\
        \specialrule{0em}{2pt}{2pt}
        The city, with its potential for both growth and decline, is explored as a potential site for the study of global economic and social trends. & The city of Afghanistan set up a temporary moratorium on violence and the UN Security Council's approval of a draft pact. About 200 people died in the country during the war, including some who were killed by soldiers.\\
        \specialrule{0em}{2pt}{2pt}
        The city, with its rich and varied histories, is discussed as a potential source of inspiration and lessons for global governance and cooperation. & The city of London rocked the world today as an international panel investigating the deaths of two Palestinians in Iraq presented a stunning array of medical services for both sides.\\
        
        \hline
        
        \hline
    \end{tabular}
    \caption{ChatGPT's cases for world topic control.}
    \label{tab:casecompare_world}
\end{table*}
\section{Cases Study}
We demonstrate generated sentences of single-attribute control and multi-attribute control in Table \ref{tab:case_single} and Table \ref{tab:case_multi}, respectively.

\begin{table*}
    \centering
    \begin{tabular}{|p{1.6cm}||p{13.4cm}|}
        \hline
        
        \hline
        \textbf{Attribute} & \textbf{PriorControl} \\
        \hline
        \hline
        Negative & Furthermore, this movie is a complete waste of time and money. It makes sense to see an effort at making a memorable plot and characters but it doesn't get any better than that. The script is riddled with holes and the direction was crudely edited.\\
        \specialrule{0em}{2pt}{2pt}
        Positive & In summary, this movie is very well written. One of the best movies ever made. This movie is filled with comedy and love. The cast is fantastic, especially those who love one particular piece of popular mythology.\\
        \specialrule{0em}{2pt}{2pt}
        World & Foundational to this is a promise of enhanced security, the U.S. military said Monday. The U.S. military has sought a chance to free up some Palestinian leader-elect Omar Fatman's loose agent.\\
        \specialrule{0em}{2pt}{2pt}
        Sports & More importantly, Atlanta United will continue to stand behind its record-breaking double-header. The United States has been through a difficult season with a disastrous effort at the Stadium of Champions.\\
        \specialrule{0em}{2pt}{2pt}
        Business & An illustration of how the world economy fell in 2003, another year after an unexpected surge in solar energy, suggests the primary driver of global corporate cash reserves is unlikely to be investor confidence alone.\\
        \specialrule{0em}{2pt}{2pt}
        Sci/Tech & The country's micro-phone technology is designed for use in conjunction with Apple Computer, a new source said. The technology used in conjunction with the open-source software Sericon OS was originally designed for cell phones.\\
        \specialrule{0em}{2pt}{2pt}
        NonToxic & The book is not an archive, but rather a revision to the article itself, so why don't you join discussions with the contributor to see where they're supposed references?\\
        
        \hline
        
        \hline
    \end{tabular}
    \caption{PriorControl's cases for single-attribute control.}
    \label{tab:case_single}
\end{table*}

\begin{table*}
    \centering
    \begin{tabular}{|p{1.6cm}||p{13.4cm}|}
        \hline
        
        \hline
        \textbf{Attribute} & \textbf{PriorControl} \\
        \hline
        \hline
        \textcolor{sred}{Negative} \textcolor{sblue}{World}\; NonToxic & This essay discusses a \textcolor{sred}{troubling situation} involving \textcolor{sblue}{an Islamic leader} and his associates in \textcolor{sblue}{the United States}. The issue is highlighted by the fact that \textcolor{sblue}{President Bush} \textcolor{sred}{was forced to resign} on Friday after repeated attempts \textcolor{sred}{failed to} convince him of the importance of keeping track of people.\\
        \specialrule{0em}{2pt}{2pt}
        \textcolor{sred}{Negative} \textcolor{sblue}{Sports}\; NonToxic & The relationship between Mike and Tony Duke is one of \textcolor{sred}{the worst ones} in \textcolor{sblue}{baseball}. Not only did he have \textcolor{sred}{a bad shot} at the \textcolor{sblue}{championship}, but there were signs of him being involved in \textcolor{sred}{some serious dispute}.\\
        \specialrule{0em}{2pt}{2pt}
        \textcolor{sred}{Negative} \textcolor{sblue}{Business} NonToxic & \textcolor{sblue}{The chicken industry} is \textcolor{sred}{turning out to be a very different product than it previously thought}, according to the \textcolor{sblue}{US Treasury Department}. In an indication of how much \textcolor{sblue}{money} they are \textcolor{sred}{losing}, \textcolor{sblue}{Federal Reserve officials} announced Wednesday that its quarterly \textcolor{sred}{\textcolor{sblue}{profit} was zero}.\\
        \specialrule{0em}{2pt}{2pt}
        \textcolor{sred}{Negative} \textcolor{sblue}{Sci/Tech} NonToxic & Once upon a time I thought this would be an interesting article in the \textcolor{sblue}{website}. \textcolor{sred}{Unfortunately, it is not.}  As I mentioned above, the source is \textcolor{sred}{not credible} and the editing is \textcolor{sred}{extremely sloppy}.\\
        \specialrule{0em}{2pt}{2pt}
        \textcolor{sred}{Positive} \textcolor{sblue}{World}\; NonToxic & The issue focused on the possibility of establishing \textcolor{sred}{a permanent peace settlement} in \textcolor{sblue}{Iraq}, an indication that \textcolor{sblue}{President Bush} is \textcolor{sred}{keen} to do it.\\
        \specialrule{0em}{2pt}{2pt}
        \textcolor{sred}{Positive} \textcolor{sblue}{Sports}\; NonToxic & The road ahead of the \textcolor{sblue}{Olympic Games} is steep, but David Beckham has been in a mood for more serious thinking since last week. The man who has made \textcolor{sred}{famous} his own love story and whose career is \textcolor{sred}{dominated by explosions}, \textcolor{sred}{smiles} at all the other \textcolor{sblue}{players}.\\
        \specialrule{0em}{2pt}{2pt}
        \textcolor{sred}{Positive} \textcolor{sblue}{Business} NonToxic & The potato \textcolor{sblue}{industry} has \textcolor{sred}{gained a foothold} in the United States as well as elsewhere, according to a new report. The \textcolor{sblue}{company} is expected to \textcolor{sred}{bring up} its share of \textcolor{sblue}{world marketplaces} after long struggling to \textcolor{sred}{reach higher levels}.\\
        \specialrule{0em}{2pt}{2pt}
        \textcolor{sred}{Positive} \textcolor{sblue}{Sci/Tech} NonToxic & To conclude, this film is an \textcolor{sred}{excellent} example of how a \textcolor{sblue}{modern internet} user can become a \textcolor{sred}{great} person and \textcolor{sred}{enjoy} the world's \textcolor{sred}{greatest} \textcolor{sblue}{television show}. In fact, it is quite \textcolor{sred}{remarkable} to see someone in a different life than you are.\\
        \hline
        
        \hline
    \end{tabular}
    \caption{PriorControl's cases for multi-attribute control.}
    \label{tab:case_multi}
\end{table*}

\end{document}